\documentclass[10pt,journal,compsoc]{IEEEtran}

\IEEEoverridecommandlockouts
\usepackage{amssymb}
\usepackage{algorithm}
\usepackage{algorithmic}
\usepackage{bibentry}
\usepackage{amsfonts}
\usepackage{subfigure}
\usepackage{multirow}
\usepackage{amsmath}
\usepackage{graphicx}
\usepackage{epsfig}
\usepackage{color}
\usepackage{epstopdf}
\usepackage{rotating}
\usepackage{caption}
\usepackage{float}
\usepackage{multicol}
\usepackage{amssymb}
\usepackage{graphicx}
\usepackage{bbding}
\usepackage{algorithm}
\usepackage{algorithmic}
\usepackage{balance}
\usepackage{microtype}
\usepackage{graphicx}
\usepackage{subfigure}
\usepackage{booktabs} 
\usepackage{multirow}
\usepackage{amsmath}
\usepackage{makecell}
\usepackage{mathtools}
\usepackage{etoolbox}
\usepackage{cases}
\usepackage{enumitem}
\usepackage{xcolor}
\usepackage{subfigure}

\usepackage{tabularx}
\usepackage{threeparttable}
\usepackage{soul}
\usepackage{color}
\newcommand{\red}[1]{\textcolor{red}{#1}}
\newcommand{\blue}[1]{\textcolor{blue}{#1}}
\newcommand{\green}[1]{\textcolor{green}{#1}}
\definecolor{brown}{RGB}{139,64,0}
\definecolor{pink}{RGB}{255,170,182}
\definecolor{purple}{RGB}{160,32,240}

\usepackage{amssymb}

\usepackage{cases}
 
\usepackage{cite}
\usepackage{amsmath,amssymb,amsfonts}
\usepackage{algorithmic}
\usepackage{graphicx}
\usepackage{textcomp}
\usepackage{xcolor}
\def\BibTeX{{\rm B\kern-.05em{\sc i\kern-.025em b}\kern-.08em
    T\kern-.1667em\lower.7ex\hbox{E}\kern-.125emX}}
    
\usepackage{mathtools}

\usepackage{amssymb}
\usepackage{amsthm}
\usepackage{color}
\usepackage{multirow}

\newcommand{\cut}[1]{{}}

\makeatletter
\let\@@span\span
\def\sp@n{\@@span\omit\advance\@multicnt\m@ne}
\makeatother

\newcommand{\bc}{\begin{center}}
\newcommand{\ec}{\end{center}}

\newcommand{\bdm}{\begin{displaymath}}
\newcommand{\edm}{\end{displaymath}}

\newcommand{\beq}{\begin{equation}}
\newcommand{\eeq}{\end{equation}}

\newcommand{\bfl}{\begin{flushleft}}
\newcommand{\efl}{\end{flushleft}}

\newcommand{\bt}{\begin{tabbing}}
\newcommand{\et}{\end{tabbing}}

\newcommand{\beqn}{\begin{align}}
\newcommand{\eeqn}{\end{align}}

\newcommand{\beqs}{\begin{align*}} 
\newcommand{\eeqs}{\end{align*}}  

\begin{document}
\title{Empowering Molecule Discovery for Molecule-Caption Translation with Large Language Models: A ChatGPT Perspective}








\author{Jiatong Li, Yunqing Liu, Wenqi Fan, Xiao-Yong Wei, 
Hui Liu, Jiliang Tang, and Qing Li

\IEEEcompsocitemizethanks{
\IEEEcompsocthanksitem J. Li, Y. Liu, W. Fan, X. Wei, and Q. Li are with the Department of Computing, The Hong Kong Polytechnic University. E-mail:  \{jiatong.li, yunqing617.liu\}@connect.polyu.hk, wenqifan03@gmail.com, x1wei@polyu.edu.hk, csqli@comp.polyu.edu.hk. 
\IEEEcompsocthanksitem J. Tang and H. LIu are with Michigan State University. E-mail: tangjili@msu.edu and liuhui7@msu.edu.

}
\thanks{(Corresponding authors: Yunqing Liu and Qing Li.)}
}

\markboth{IEEE TRANSACTIONS ON KNOWLEDGE AND DATA ENGINEERING, SUBMISSION 2023}%
{Shell \MakeLowercase{\textit{et al.}}: Bare Demo of IEEEtran.cls for Computer Society Journals}

\IEEEtitleabstractindextext{%

\begin{abstract}
Molecule discovery plays a crucial role in various scientific fields, advancing the design of tailored materials and drugs, which contributes to the development of society and human well-being. 
Specifically, molecule-caption translation is an important task for molecule discovery, aligning human understanding with molecular space.
However, most of the existing methods heavily rely on domain experts, require excessive computational cost, or suffer from sub-optimal performance. 
On the other hand, Large Language Models (LLMs), like ChatGPT, have shown remarkable performance in various cross-modal tasks due to their powerful capabilities in natural language understanding, generalization, and in-context learning (ICL), which provides unprecedented opportunities to advance molecule discovery.
Despite several previous works trying to apply LLMs in this task, the lack of domain-specific corpus and difficulties in training specialized LLMs still remain challenges.
In this work, we propose a novel LLM-based framework (\textbf{MolReGPT}) for molecule-caption translation, where an In-Context Few-Shot Molecule Learning paradigm is introduced to empower molecule discovery with LLMs like ChatGPT to perform their in-context learning capability without domain-specific pre-training and fine-tuning.
MolReGPT leverages the principle of molecular similarity to retrieve similar molecules and their text descriptions from a local database to enable LLMs to learn the task knowledge from context examples. 
We evaluate the effectiveness of MolReGPT on molecule-caption translation, including molecule understanding and text-based molecule generation. Experimental results show that compared to fine-tuned models, MolReGPT outperforms MolT5-base and is comparable to MolT5-large without additional training. 
To the best of our knowledge, MolReGPT is the first work to leverage LLMs via in-context learning in molecule-caption translation for advancing molecule discovery. Our work expands the scope of LLM applications, as well as providing a new paradigm for molecule discovery and design.
Notably, our implementation is available at: \url{https://github.com/phenixace/MolReGPT}
\end{abstract}

}

\maketitle
\IEEEpeerreviewmaketitle

 \section{Introduction}

\begin{figure}[t]
\centering

\begin{minipage}[t]{4.0cm}
\vspace{0pt}
\subfigure[Molecule Representations.]{ 
\includegraphics[scale=0.42]{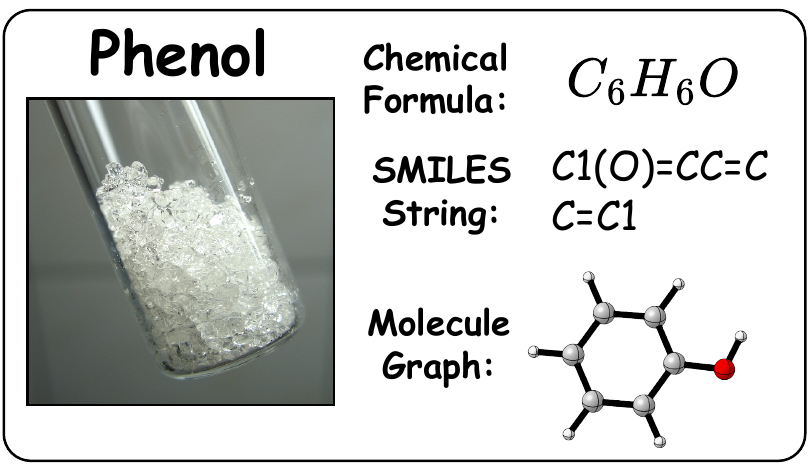}
}
\vspace{0.3cm}
\subfigure[Molecule Captioning.]
{ 
\includegraphics[scale=0.42]{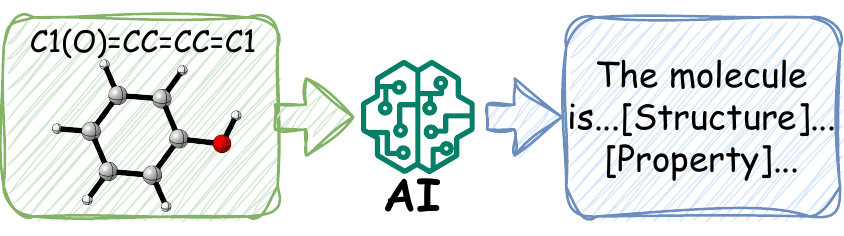}
}
\vspace{0.3cm}
\subfigure[Text-based Molecule Generation.]
{   
\includegraphics[scale=0.42]{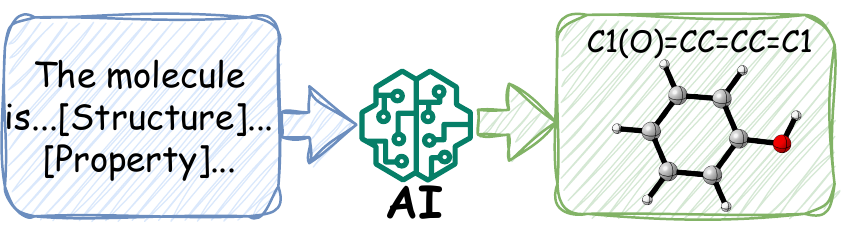}
}
\end{minipage}
\subfigure[Empowering ChatGPT with molecule captioning and text-based molecule generation abilities.]{ 
\begin{minipage}[t]{4.5cm} 
\centering 
\vspace{0pt}
\includegraphics[width=\textwidth]{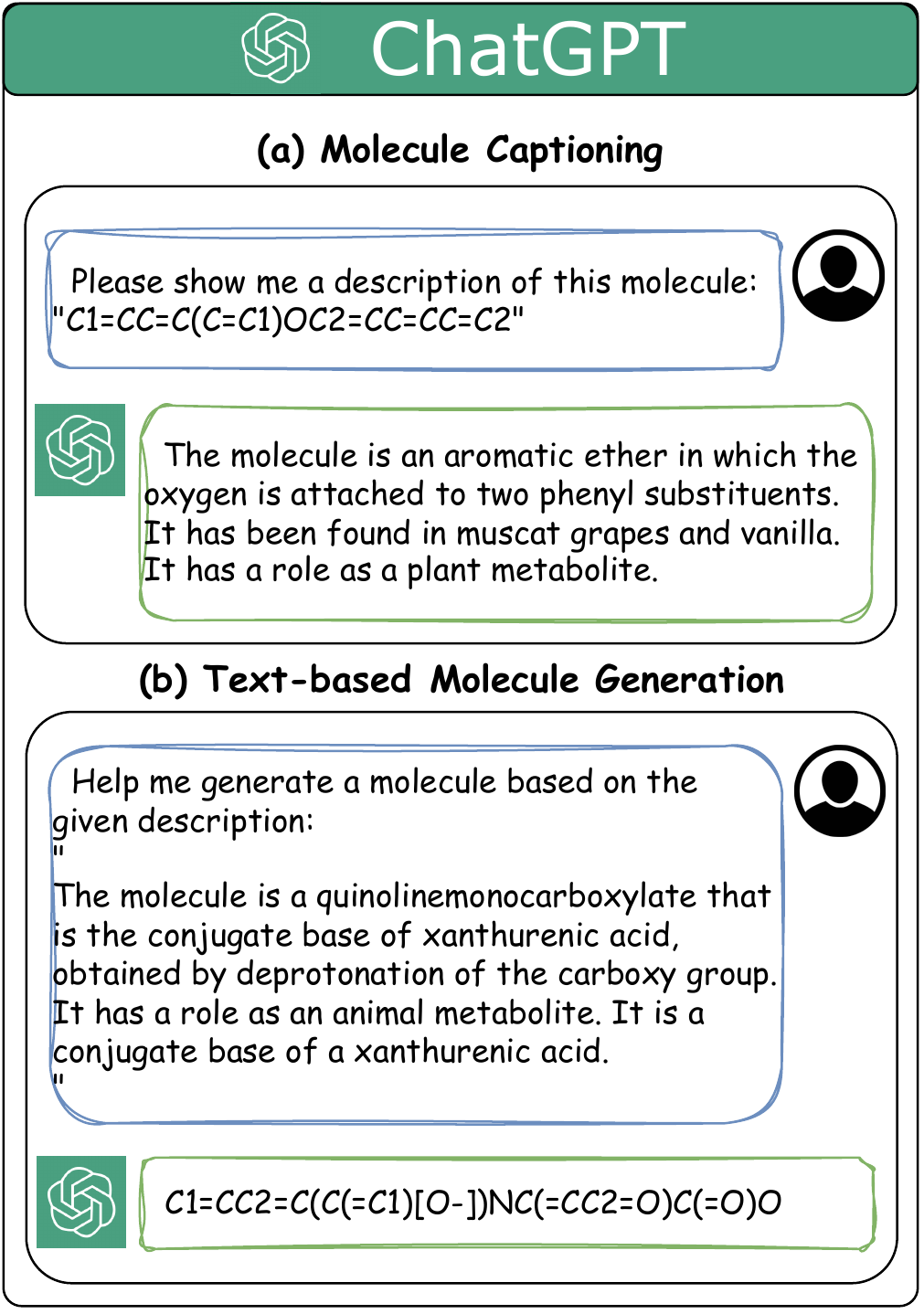}
\end{minipage}
}

\caption{An illustration of molecule-caption translation. (a) Representations of a molecule.
(b) Molecule captioning (Mol2Cap) aims to generate a text caption describing the features of the molecule. 
(c) Text-based molecule generation (Cap2Mol) generates a corresponding molecule to the given caption.
(d) Large language models (e.g., ChatGPT) can perform Mol2Cap and Cap2Mol with well-designed prompts.}
\label{fig:bot} 

\end{figure}

As the foundation of chemical compounds, molecules are composed of two or more atoms that are chemically bonded together, denoting their unique chemical properties dictated by their specific structures \cite{xu2023plasma}.
With a comprehensive understanding of molecules, scientists can effectively design materials, drugs, and products with tailored characteristics and functionalities, impacting a variety of crucial fields such as chemistry \cite{weng2021late}, pharmacology \cite{ding2019selective}, and material science \cite{higuchi2023material}.  

Recently, computational technologies such as artificial intelligence (AI) have emerged as powerful tools to expedite the discovery of new molecules \cite{urbina2022commoditization}. 
Specifically, molecules can be represented as simplified molecular-input line-entry system (\textbf{SMILES}) strings \cite{weininger1988smiles}, illustrated in Figure \ref{fig:bot} (a), which can be effectively processed by deep sequence models like Recurrent Neural Networks \cite{arus2019randomized} and Transformers \cite{honda2019smiles}.
These AI-powered models enable researchers to understand molecular properties and functionalities and thus create promising compounds in a more efficient and cost-effective manner. 
For example, in order to generate new molecules and better comprehend them, a novel task that translates between molecules and natural language has been proposed by using language models like Text2Mol ~\cite{edwards2021text2mol} and MolT5 ~\cite{edwards-etal-2022-translation}. It consists of two sub-tasks: molecule captioning (Mol2Cap) and text-based molecule generation (Cap2Mol). 
As shown in Figure~\ref{fig:bot} (b-c), the goal of Mol2Cap is to generate a text caption describing molecule features (e.g. structures and characteristics). Specifically, the text caption first demonstrates the molecule structure by describing functional group positions and the according IUPAC name. Then, the characteristics of the molecule will be discussed, including the family that it belongs to and the chemical features that affect its practical use.
On the other hand, Cap2Mol aims to generate the corresponding molecule (i.e., SMILES string) based on the given text caption, where the structure and feature information of the molecules could be used to infer the SMILES representation of the molecule.
For example, given the molecule \emph{"CCCCCCCCCCCCCCCCCCCCCCCCCCCCC"}, its caption writes \emph{"The molecule is a straight-chain alkane comprising of 29 carbon atoms. It has a role as a plant metabolite and a volatile oil component"}, where \emph{"straight-chain alkane"} and \emph{"29 carbon atoms"} directly describe the molecule structure, while the last sentence shows the functions of the molecule.

Despite the impressive progress that has been made in the molecule-caption translation task~\cite{edwards2021text2mol, edwards-etal-2022-translation, su2022molecular}, the majority of existing advanced approaches still suffer from several limitations.
\emph{First}, the design of model architectures heavily relies on the labour of domain experts, which can significantly limit the development of AI-powered molecule discovery. 
\emph{Second}, most existing methods follow the \emph{pre-train\&fine-tuning} paradigm for molecule-caption translation, putting demanding requirements on computational and domain resources.
\emph{Third}, existing approaches such as Text2Mol and MolT5 fall short in their inability to generalize to unseen examples. Therefore, it is desired to design a novel paradigm for molecule-caption translation.

Recently, Large Language Models (\textbf{LLMs}), scaling up their weights to the billion level, have achieved tremendous success not only in the field of Natural Language Processing (NLP) but also in some cross-modal areas like computer vision \cite{zhu2023minigpt}, recommender systems \cite{bao2023tallrec}, and molecule discovery~\cite{edwards-etal-2022-translation}. 
Meanwhile, in addition to the impressive capabilities in natural language understanding and generation, 
LLMs also demonstrate their powerful generalization and reasoning capabilities \cite{rubin2022learning, min2022metaicl}, which can generalize to other unseen tasks via in-context learning without the necessity of being fine-tuned, largely reducing computational cost. 
Therefore, LLMs provide unprecedented potential to advance molecule discovery, specifically the task of molecule-caption translation.

Although building specific LLMs in molecule discovery has immense potential for advancing scientific research, we also face significant challenges. 
\emph{First}, due to privacy and security concerns, many advanced large language models (e.g., GPT-3.5 and GPT-4) are not publicly available, where LLMs' architectures and parameters are not released publicly for fine-tuning in downstream tasks.
\emph{Second}, owing to their complex architectures and the extensive data required, training advanced LLMs requires significant computing resources and domain corpus, leading to high costs and substantial energy consumption. 
For instance, it has been reported that the cost of \emph{one single training session} for GPT-3 exceeds 1 million US dollars.
As a result, it is very challenging for us to re-design our own LLMs with pre-training and fine-tuning in specific downstream tasks. 
\emph{At last}, it still lacks proper guidelines/paradigms for scientific researchers to make use of powerful LLMs like ChatGPT to enhance their own study in molecule discovery.

To address such challenges, as the early exploration attempt to take advantage of the powerful capabilities of LLMs in the molecule discovery field, in this work, we propose a novel solution to teach LLMs with specific in-context examples for translating between molecules and natural language, as illustrated in Figure~\ref{fig:bot} (d).
More specifically, inspired by the latest ChatGPT, a retrieval-based In-Context Few-Shot Molecule Learning paradigm is developed to conduct two sub-tasks (i.e., Mol2Cap and Cap2Mol) without fine-tuning the LLMs, where n similar molecule-caption pairs are retrieved as context instances under the guidance of molecular similarity, including BM25-based caption retrieval and Morgan Fingerprints-based molecule retrieval. 
Experiments show that MolReGPT (GPT-4-0314) achieves Text2Mol scores of 0.585 in Mol2Cap and 0.593 in Cap2Mol, showing comparable performance to MolT5-large in Mol2Cap and even outperforming MolT5-large in Cap2Mol without any fine-tuning steps, increasing the Text2Mol metric by 0.5\% and 6\%, respectively. 

Our major contributions are summarized as follows:
\begin{itemize}[leftmargin=*] 
    \item We introduce a principle strategy based on LLMs, In-Context Few-Shot Molecule Learning, to perform translation between molecules and natural language for molecule discovery. To the best of our knowledge, we are the first to employ the in-context learning ability of LLMs in molecule-caption translation. Our work expands the application scope of LLMs and provides valuable insights into how these LLMs can be adapted for specific scientific tasks.

    \item We develop a novel framework (MolReGPT) to empower LLMs like ChatGPT for scientific purposes without being pre-trained or fine-tuned on domain-specific corpora. By enabling LLMs to understand and generate meaningful molecular descriptions and molecule structures, we pave the way for AI-assisted drug discovery and design. MolReGPT has the potential to accelerate the development of new pharmaceuticals and improve the efficiency of molecular research.
    
    \item We conduct comprehensive experiments on a real-world dataset with molecule-caption pairs to demonstrate the effectiveness and mechanisms of the proposed method on Mol2Cap and Cap2Mol tasks. The results show that our method could enable GPT4-0314 to achieve comparable performance to MolT5-large.

\end{itemize} 

\section{Related Work}
\label{sec:relatedwork}
 
\noindent \textbf{Molecule Discovery}. 
In recent decades, AI-powered approaches have emerged as mainstream techniques to revolutionize the process of molecule discovery~\cite{hu2023deep,fan2023generative}.
existing studies have explored advanced deep representation methods from other fields, including Convolutional Neural Network (CNN)~\cite{peng2019convolutional,le2019imotor}, Recurrent Neural Network (RNN)~\cite{grisoni2020bidirectional,amabilino2020guidelines}, and Transformer~\cite{bagal2021molgpt,wang2021multi}.
More recently, as a new task in molecule discovery, Text2Mol~\cite{edwards2021text2mol} is introduced to retrieve molecules using natural language descriptions as search queries, in which a paired dataset of molecules and their corresponding text descriptions are constructed, enabling the learning of a shared semantic embedding space for retrieval.
KV-PLM~\cite{zeng2022deep} develops a knowledgeable machine reading system pre-trained on a domain corpus, in which SMILES strings are inserted and link molecule structures with biomedical text.
What's more, a self-supervised learning framework MolT5~\cite{edwards-etal-2022-translation} is proposed to pre-train on a substantial volume of unlabeled language text and SMILES strings, enhancing the molecule-caption translation task, such as molecule captioning and text-based molecule generation.
MoMu~\cite{su2022molecular} bridges molecular graphs and natural language by pre-training molecular graphs and their semantically related text data through comparative learning.

\noindent \textbf{Large Language Models (LLMs)}. 
LLMs have been a trending topic in recent years, with numerous studies exploring their capabilities and potential applications. 
One of the most well-known LLMs is the GPT family~\cite{radford2018improving,radford2019language,ouyang2022training}, which has played a critical role in advancing the field of generative language models. As a representative of the GPT family, ChatGPT is specifically fine-tuned for conversational purposes, which can generate impressively human-like responses~\cite{openai-chatgpt}. 
In addition, other LLMs, such as LaMDA~\cite{thoppilan2022lamda}, PaLM~\cite{chowdhery2022palm}, and Vicuna~\cite{chiang2023vicuna}, also show a decent performance. 


In addition to NLP tasks, LLMs have also shown remarkable potential in various molecule discovery tasks, such as molecule understanding~\cite{bran2023chemcrow,white2023future}. For instance, ChemBERTa~\cite{chithrananda2020chemberta} leverages pre-training on an extensive corpus of chemical texts, enabling it to comprehend the structure and properties of chemical compounds. Another notable example is MoleculeSTM~\cite{liu2022multi}, which employs in-context learning in conjunction with LLMs. This approach facilitates a deeper understanding of the relationships between chemical structures and their corresponding textual descriptions. 
Furthermore, ChemGPT~\cite{frey2022neural} represents a variant of the GPT model specifically trained on chemical data. Through well-designed instructions, ChemGPT is capable of generating novel chemical structures and accurately predicting their properties.
MolT5~\cite{edwards-etal-2022-translation} shows that LLMs can perform the cross-modal transition task between molecule and text (i.e. molecule captioning task and text-based molecule generation task), which is one of the most closely related attempts to ours. 
Besides, it should be noted that MolT5 still needs to be pre-trained on domain corpus and further fine-tuned for translating between molecules and natural language, leading to huge computational costs and harsh data requirements.
\section{MolReGPT}
\label{sec:methodlogy}


\begin{figure*}[htb]

    \centering
    \includegraphics[width=\textwidth]{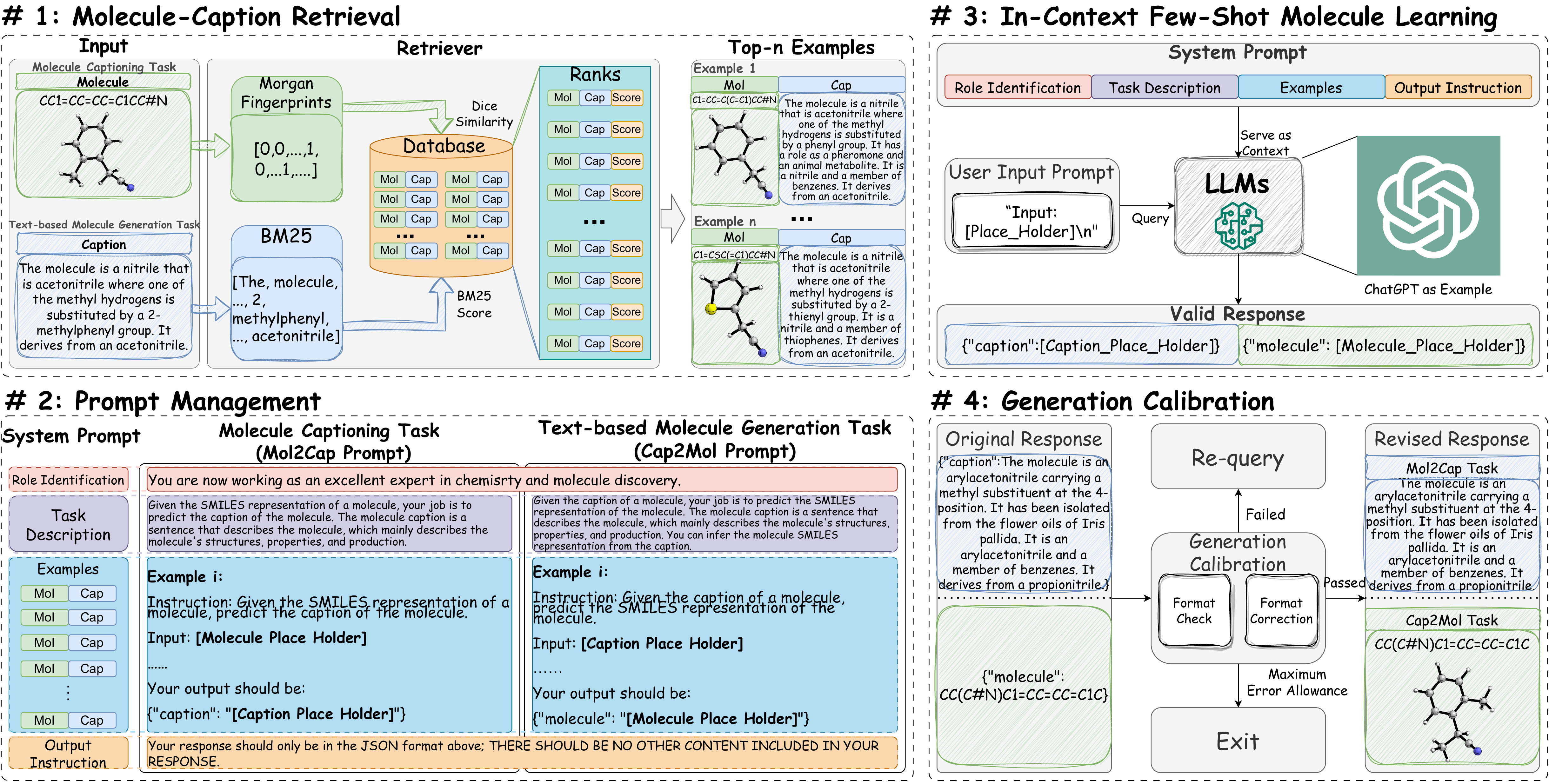}

    \caption{The workflow of MolReGPT. MolReGPT consists of four main stages. 
    In stage 1, Molecule-Caption Retrieval is employed to find \(n\) best-matched examples from the local database. 
    Then, in stage 2, Prompt Management helps construct the system prompt with the retrieved molecule-caption pairs. Following this, LLMs perform In-Context Few-Shot Molecule Learning based on the provided system prompt and user input prompt. 
    Finally, Generation Calibration ensures the desired output. 
    }
    \label{fig:prompt}

\end{figure*}


Due to the huge computation and domain data labelling costs, training or fine-tuning LLMs on the domain-specific corpus in molecule discovery is often infeasible in practice.
To address such limitations, we investigate leveraging the great capabilities of LLMs without changing the LLMs, where we propose a novel framework (\textbf{MolReGPT}) to equip ChatGPT with the ability of molecule-caption translation for molecule discovery. 
Specifically, in order to improve the quality of prediction, an In-Context Few-Shot Molecule Learning paradigm is introduced to teach ChatGPT to learn the molecule-caption translation task from context examples.
The framework of MolReGPT is shown in Figure \ref{fig:prompt}, consisting of four main stages: \emph{Molecule-Caption Retrieval, Prompt Management, In-Context Few-Shot Molecule Learning}, and \emph{Generation Calibration}, following the workflow of pre-processing, querying, and post-processing. 

\subsection{Molecule-Caption Retrieval}

In order to teach LLMs to handle the molecule-caption translation task without fine-tuning LLMs, we propose in-context few-shot molecule learning to guide LLMs to learn how to translate between molecules and text captions.
Normally, in-context learning requires $n$ random examples selected from human-annotated datasets (i.e., molecule-caption pair database), providing a general task instruction to LLMs. 
However, random examples often provide insufficient knowledge regarding the associations between natural language and molecules, as they fail to provide useful information for the detailed descriptions of functional groups and molecule features.
To mitigate this issue, we propose incorporating retrieval methods into the selection of context examples to complement the lack of task-specific knowledge in LLMs, specifically through the stage of Molecule-Caption Retrieval.
These retrieval strategies are motivated by the similar property principle, in which molecules similar in structures tend to exhibit similar characteristics ~\cite{wang2016improving}.
Namely, similar captions containing the descriptions of molecule structures and properties are used to describe similar molecules.
Therefore, via these most similar molecules or captions, we could utilize the corresponding molecule-caption pairs as context examples to guide LLMs.

Notably, the SMILES representation of molecules, as a sequence structure, can hardly reveal the actual 2-D graph topology of molecules. Hence, domain-specific methods are required for better molecular similarity calculation during the retrieval stage.
Specifically, given a SMILES string representation for Mol2Cap task, we introduce using Morgan Fingerprints (i.e., a molecular structures representation)~\cite{butina1999unsupervised} to calculate molecular similarity using Dice similarity for molecule retrieval.
Meanwhile, in Cap2Mol task, we are more focused on the details in the text captions (e.g., IUPAC names and functional group positions). Thus, BM25 caption retrieval, which is widely used in information retrieval~\cite{robertson2009probabilistic}, is proposed to compute similarity scores between captions of molecules.
In both sub-tasks, top-n molecule-caption pair examples are retrieved to serve as context examples in the system prompt.
Next, we will detail Morgan Fingerprints-based molecule retrieval and BM25-based caption retrieval.

\subsubsection{Morgan Fingerprints-based Molecule Retrieval.}  
Molecular fingerprints are numerical representations of the chemical structures of molecules, which can be used for various computational objectives~\cite{butina1999unsupervised}, such as similarity searching, property prediction, and cluster analysis.
One of the most representative molecular fingerprints is the Morgan Fingerprints (Morgan FTS). 

\begin{figure}[htb]
    \centering
    \includegraphics[width=1.0\columnwidth]{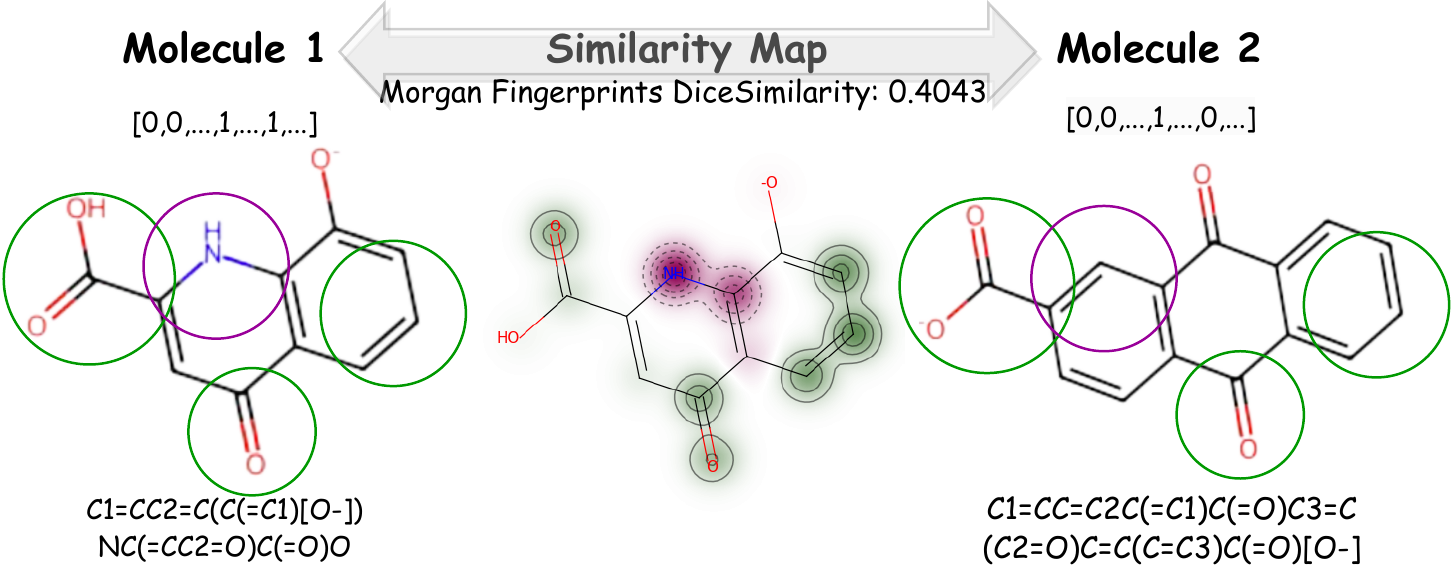}
    \caption{Illustrations of Morgan Fingerprints and Dice Similarity. 
    The two molecules will first be transformed into the Morgan Fingerprints. Then, Dice similarity will be calculated.
    The green colour corresponds to sub-structures that contribute positively to the similarity score between the molecules, 
    while the purple colour represents sub-structures that contribute negatively or have differences between the molecules. 
    }
    \label{fig:similarity}
\end{figure}

The key idea behind Morgan FTS is to capture the presence or absence of specific sub-structures or chemical fragments in a molecule. Morgan FTS follows a variant of the Morgan algorithm~\cite{butina1999unsupervised}, which encodes the structural information of a molecule by representing its connectivity patterns in a circular manner. Morgan FTS is then generated by iteratively expanding a set of atoms from a central atom in the molecule, capturing the neighbouring atoms and their bond types at each expansion step. The process continues until a pre-defined radius is reached. The result is a binary bit vector, where each bit represents the presence or absence of a particular substructure.

What's more, Morgan FTS has several advantages over other types of fingerprints, including their ability to handle molecules of varying sizes, resistance to small structural changes, and effectiveness in capturing structural similarities between molecules.
Subsequently, we apply Dice similarity~\cite{dice1945measures} to measure the similarity between the input molecule and the molecules in the local database. Mathematically, it can be expressed as:

\begin{equation}
    Dice(A, B) = \frac{(2*|A\cap B|)}{|A| + |B|},
\end{equation}

\noindent where $A$ and $B$ are the Morgan Fingerprints of two molecules. $|A|$ and $|B|$ represent the cardinality (i.e., number of sub-structures) of $A$ and $B$. $|A \cap B|$ denotes the number of sub-structures that are common to both $A$ and $B$. Dice similarity ranges from 0 to 1.
The Dice similarity is particularly useful when dealing with imbalanced datasets or focusing on the agreement between positive instances (i.e., sub-structures present in both sets) rather than the overall agreement. As shown in Figure \ref{fig:similarity}, the similarity map explains how Morgan FTS measures the similarities and differences between the two molecules.

Compared to existing molecule embedding methods, Morgan FTS together with Dice similarity provides a distinctive advantage by explicitly indicating the similarities in detailed molecular structures, as these structures are usually directly stated in the molecule captions~\cite{coupry2022application}.

\subsubsection{BM25-based Caption Retrieval.}
BM25 is one of the most representative ranking approaches in information retrieval for calculating the relevance of the documents to the given query. 
The idea is based on the term frequency-inverse document frequency (TF-IDF), which measures how often a term appears in a document (i.e., caption) and how rare it is in the corpus of documents (i.e., the local database) ~\cite{aizawa2003information}. In addition, BM25 considers the caption's length and the position of the query terms in the caption.

In the Cap2Mol task, we use the input caption as the query sentence, while the captions in the local database (i.e., the training set), are served as the corpus of documents, where each caption represents a document.
Mathematically, the formula of BM25 can be defined as follow:
{\small
\begin{align}
    score(Q,\!D)\!=\!\sum_{i=1}^N\!I\!D\!F(\!q_i\!)\!*\!\frac{f(q_i,D)*(k_1+1)}{f(\!q_i,\!D\!)\!+\!k_1\!*\!(\!1\!-\!b\!+\!b\!*\!\frac{|D|}{avgdl}\!)},
\end{align}
}
\noindent where $D$ is the caption corpus and $Q$ is the query caption. $N$ is the number of query terms in the query caption, $q_i$ is the $i$-th query term, $IDF(q_i)$ is the inverse document frequency of $q_i$, $f(q_i, D)$ is the term frequency of $q_i$ in $D$, $k_1$ and $b$ are tuning parameters, $|D|$ is the length of $D$, and $avgdl$ is the average caption length in the corpus.
In Cap2Mol task, as we discussed, structure details usually bring more information gain to the prediction of the molecule SMILES representations, while such details are difficult to capture at the semantic level.
In this case, BM25 caption retrieval is applied to calculate the similarity scores between captions at the token level so that the relevant molecule structures described by captions can be learnt via molecule-caption pairs.

\subsection{Prompt Management}
System prompts and user input prompts are two important parts to enable the in-context learning ability of LLMs. User prompts are usually more complex and contain essential instructions for task solving and format formalization, where user prompts are defined to formalize the user inputs. To help LLMs understand the task and generate desired outputs, Prompt Management is proposed to design the system prompt templates, which are further completed with the context examples. 
As shown in stage 2 of Figure \ref{fig:prompt}, the system prompts consist of the following four parts:
\begin{itemize}[leftmargin=*] 
    \item   \textbf{Role Identification} aims to help LLMs identify the role of experts in the chemistry and molecule discovery domain.
By establishing this role, LLMs are encouraged to generate responses aligning with the expected expertise.

    \item \textbf{Task Description} provides a comprehensive explanation of the task's content, ensuring that LLMs have a clear understanding of the specific task they need to address.
It also includes critical definitions to clarify terms or concepts that are specialized in the molecule-caption translation task.

     \item 
\textbf{Context Examples} serves as the evidence for the molecule-caption translation task, allowing LLMs to leverage the information contained within the molecule-caption pairs via in-context learning to generate better responses.

\item \textbf{Output Instruction} specifies the desired format for the response. Here, we restrict the output to a JSON format for further processing or analysis.
\end{itemize}


\subsection{In-Context Few-Shot Molecule Learning}

Recently, as an alternative to fine-tuning, in-context learning provides great opportunities to teach LLMs to make predictions based on a few context examples.
In this work, we introduce In-Context Few-Shot Molecule Learning to perform the Mol2Cap and Cap2Mol tasks without fine-tuning LLMs and analyse the underlying mechanism.
This stage is to utilize both the system prompt and user input prompt to query the LLMs.
In particular, the combination of the system prompt and user input prompt provides LLMs with a clear guideline via in-context learning; the system prompt establishes domain expertise for molecule-caption translation, while the user prompt narrows the focus and directs the model's attention to the specific user input. 
As a result, LLMs can learn how to perform the molecule-caption translation from the given task context, without the necessity to modify their parameters.

The formulas below describe the differences between fine-tuning, prompting, and In-context Few-Shot Molecule Learning.
Let $M$ be the large language model, $m$ be the molecule, $c$ be the corresponding molecule caption, and $\theta$ be the parameters of the LLM.

\noindent\textbf{Fine-tuning} process for the Mol2Cap and Cap2Mol task aims to help LLMs learn the translation probability between the molecule and its text caption: 

\begin{equation}
    c = L(m;\theta^*_m), 
\end{equation}
\begin{equation}
    m = L(c;\theta^{*}_c),
\end{equation}

\noindent where $\theta^*_m$ and $\theta^{*}_c$ are the updated parameters after being fine-tuned on the entire training set ($\theta^*_m$ for Mol2Cap and $\theta^*_c$ for Cap2Mol). 

\noindent\textbf{Prompting} uses prompts to wrap the inputs and guide the generation of LLMs without changing the parameters. It can be defined as:

\begin{equation}
    c = L(p_m(m);\theta),
\end{equation}
\begin{equation}
    m = L(p_c(c);\theta),
\end{equation}

\noindent where $p_m(\cdot)$ and $p_c(\cdot)$ are the Prompt Management templates that transform the original user input (molecules $p_m$ or captions $p_c$) into system prompts with the user input prompts for querying LLMs, and $\theta$ is the original parameters without being fine-tuned.

\noindent\textbf{In-Context Few-Shot Molecule Learning} targets the alignment between text and molecule structures. Our method can be formulated as:
\begin{equation}
    c = L(p_m(m)||T_{m \rightarrow c};\theta),
\end{equation}
\begin{equation}
    m = L(p_c(c)||T_{c \rightarrow m};\theta),
\end{equation}

where $T_{m \rightarrow c} = (m_1 \rightarrow c_1)||(m_2 \rightarrow c_2)|| ... ||(m_n \rightarrow c_n)$ and $T_{c \rightarrow m} = (c_1 \rightarrow m_1)||(c_2 \rightarrow m_2)|| ... ||(c_n \rightarrow m_n)$ are the retrieved context examples for the Mol2Cap and Cap2Mol task. Here, $(c_i \rightarrow m_i)$ denotes the $i$-th retrieved example that illustrates how to transfer from caption $c_i$ to the molecule $m_i$, while $(m_i \rightarrow c_i)$ represents the alignment from the molecule $m_i$ to its caption $c_i$, where $i \in [1,n]$. Notably, $\theta$ is still the original parameter of LLMs without any modification.

Through the way of learning the molecule-caption translation from context examples, LLMs can grasp the alignment between text description and the molecule structures it implies. Compared to fine-tuning, our method does not require additional model training. Compared to prompting, our method is more explainable and robust due to the in-context learning process.

\subsection{Generation Calibration}
Despite specifying the desired output format, LLMs (e.g., ChatGPT) can occasionally produce unexpected responses, including incorrect output formats and denial of answering. 
To address these issues, a generation calibration mechanism is introduced to validate the response from LLMs.

In Generation Calibration, we first check the format of original responses. 
If the format is not correct,
several pre-defined format correction strategies, such as Regular Matching, are introduced to correct the format and extract the desired output from the response. 
If the original response passes the format check or can be calibrated, it is considered valid and accepted as a final response. 
However, if the original response fails the format check and cannot be corrected within the predefined strategies, we initiate re-queries. 
Notably, there is a special case for re-queries. When the original response reports the ``Exceed Maximum Input Length Limitation" error, we will remove the longest example in the re-query phase until the query length meets the length limitation.
The re-query process involves making additional queries to the LLMs until a valid response is obtained or until the maximum error allowance is reached to ensure that the system does not get stuck in an endless loop.


\section{Experiment}
\label{sec:Experiments}

\begin{table*}[htb]
    \centering
    \resizebox{2.0\columnwidth}{!}{
    \begin{tabular}{c|c|c|c|c|c|c|c}
    \toprule
    Methods & BLEU-2$\uparrow$ & BLEU-4$\uparrow$ & ROUGE-1$\uparrow$ & ROUGE-2$\uparrow$ & ROUGE-L$\uparrow$ & METEOR$\uparrow$ & Text2Mol$\uparrow$ \\
    \midrule
    Transformer \cite{edwards-etal-2022-translation} & 0.061 & 0.027 & 0.204 & 0.087 & 0.186 & 0.114 & 0.057 \\
    GPT-3.5-turbo (zero-shot) & 0.103 & 0.050 & 0.261 & 0.088 & 0.204 & 0.161 & 0.352 \\ 
    LLama-2-7B (zero-shot) & 0.094 & 0.039 & 0.169 & 0.054 & 0.142 & 0.175 & 0.153\\
    LLama-2-7B (2-shot MolReGPT) & 0.489 & 0.409 & 0.535 & 0.374 & 0.472 & 0.495 & 0.466\\
    \midrule
    T5-base \cite{edwards-etal-2022-translation} & 0.511 & 0.423 & 0.607 & 0.451 & 0.550 & 0.539 & 0.523 \\ 
    MolT5-base \cite{edwards-etal-2022-translation} & 0.540 & 0.457 & \underline{0.634} & \underline{0.485} & \underline{0.578} & 0.569 & 0.547 \\ 
    GPT-3.5-turbo (10-shot MolReGPT) & 0.565 & 0.482 & 0.623 & 0.450 & 0.543 & 0.585 & 0.560 \\
    \midrule
    T5-large \cite{edwards-etal-2022-translation} & 0.558 & 0.467 & 0.630 & 0.478 & 0.569 & 0.586 & 0.563 \\
    MolT5-large \cite{edwards-etal-2022-translation} & \underline{0.594} & \underline{0.508} & \textbf{0.654} & \textbf{0.510} & \textbf{0.594} & \textbf{0.614} & \underline{0.582} \\
    GPT-4-0314 (10-shot MolReGPT) & \textbf{0.607} & \textbf{0.525} & \underline{0.634} & 0.476 & 0.562 & \underline{0.610} & \textbf{0.585} \\
    \bottomrule
    \end{tabular}
            
    }

    \caption{The performance of Mol2Cap on ChEBI-20. Experimental results for Transformer, T5-base, MolT5-base, T5-large, and MolT5-large are retrieved from ~\cite{edwards-etal-2022-translation}. Due to the input length limitation, we apply 2-shot MolReGPT to llama-2-7B and 10-shot MolReGPT to GPT models. The \textbf{best} scores are in bold, and the \underline{second-best} scores are underlined.}

    \label{tab:m2c}
\end{table*}

\begin{table*}[htb]
    \centering      
    \resizebox{2.0\columnwidth}{!}{
    \begin{tabular}{c|c|c|c|c|c|c|c|c|c}
    \toprule
    Method & BLEU$\uparrow$ & EM$\uparrow$ & Levenshtein$\downarrow$ & MACCS FTS$\uparrow$ & RDK FTS$\uparrow$ & Morgan FTS$\uparrow$ & FCD$\downarrow$ & Text2Mol$\uparrow$ & Validity$\uparrow$ \\
    \midrule
    Transformer \cite{edwards-etal-2022-translation} & 0.499 & 0.000 & 57.66 & 0.480 & 0.320 & 0.217 & 11.32 & 0.277 & \textbf{0.906} \\ 
    GPT-3.5-turbo (zero-shot) & 0.489 & 0.019 & 52.13 & 0.705 & 0.462 & 0.367 & 2.05 & 0.479  & 0.802 \\ 
    LLama-2-7B (zero-shot) & 0.104 & 0.000 & 84.18 & 0.243 & 0.119 & 0.089 & 42.01 & 0.148 & 0.631\\
    LLama-2-7B (2-shot MolReGPT) & 0.693 & 0.022 & 36.77 & 0.808 & 0.717 & 0.609 & 4.90 & 0.149 & 0.761 \\
    \midrule
    T5-base \cite{edwards-etal-2022-translation} & 0.762 & 0.069 & 24.950 & 0.731 & 0.605 & 0.545 & 2.48 & 0.499 & 0.660 \\ 
    MolT5-base \cite{edwards-etal-2022-translation} & 0.769 & 0.081 & 24.458 & 0.721 & 0.588 & 0.529 & 2.18 & 0.496 & 0.772 \\
    GPT-3.5-turbo (10-shot MolReGPT) & 0.790 & 0.139 & 24.91 & \underline{0.847} & 0.708 & 0.624 & \underline{0.57} & \underline{0.571} & 0.887 \\ 
    \midrule
    T5-large \cite{edwards-etal-2022-translation} & \underline{0.854} & 0.279 & \underline{16.721} & 0.823 & 0.731 & 0.670 & 1.22 & 0.552 & 0.902 \\
    MolT5-large \cite{edwards-etal-2022-translation} & \underline{0.854} & \textbf{0.311} & \textbf{16.071} & 0.834 & \underline{0.746} & \underline{0.684} & 1.20 & 0.554 & \underline{0.905} \\
    GPT-4-0314 (10-shot MolReGPT) & \textbf{0.857} & \underline{0.280} & 17.14 & \textbf{0.903} & \textbf{0.805} & \textbf{0.739} & \textbf{0.41} & \textbf{0.593} & 0.899 \\
    \bottomrule
    \end{tabular}
    }

    \caption{Cap2Mol results on ChEBI-20. Experimental results for Transformer, T5-base, MolT5-base, T5-large, and MolT5-large are retrieved from \cite{edwards-etal-2022-translation}. Due to the input length limitation, we apply 2-shot MolReGPT to llama-2-7B and 10-shot MolReGPT to GPT models. The \textbf{best} scores are in bold, and the \underline{second-best} scores are underlined.}

    \label{tab:c2m}

\end{table*}

In this section, we aim to evaluate the effectiveness of our proposed MolReGPT by conducting comprehensive experiments on molecule-caption translation task. 

\subsection{Experimental Settings}
We first introduce the basic experimental settings. In this work, we use ChatGPT through the OpenAI API\footnote{https://openai.com/blog/openai-api} with backend model \textbf{GPT-3.5-turbo} and \textbf{GPT-4-0314}, which can not be fine-tuned in our tasks. To assure the model agnosticism of MolReGPT, we also apply an extra open-source LLM, \textbf{Llama-2-7b-chat-hf} (i.e., Llama-2-7B) \cite{touvron2023llama}, for comparison. Notably, due to the input length limitation and hardware limitation, we could only apply 2-shot MolReGPT to this model.
Besides, we will provide an overview of the data and metrics employed in this section.

\noindent\textbf{Dataset.}
The research on molecule-caption translation is still in the early stage, and there is only one public dataset ChEBI-20 \cite{edwards2021text2mol}, which contains 33,010 molecule-caption pairs. To ensure consistency, we adhere to the data split process as used in MolT5 ~\cite{edwards-etal-2022-translation}, dividing the dataset into 80/10/10\% train/validation/test splits. 
For our method evaluation, we focus on the test split while utilizing the training set as the local database to retrieve n-shot examples for in-context learning. 

\noindent\textbf{Evaluation Metrics.}
In terms of evaluation metrics, we align with the metrics adopted in MolT5 ~\cite{edwards-etal-2022-translation}. By adopting these metrics, we ensure consistency and enable a fair assessment of the performance of our method.

\noindent\textbf{Baselines.}
Specifically, the following baselines are selected for performance evaluation. Most of these baselines are only fine-tuned on ChEBI-20 because of their limited maximum input length and poor reasoning capabilities. 
\begin{itemize} [leftmargin=*] 
    \item \textbf{Transformer}~\cite{vaswani2017attention}. This method is the most representative language architecture to process natural language. 
    A vanilla Transformer model with six encoder and decoder layers, directly trained on ChEBI-20. Notably, this model is not pre-trained. 
    
    \item \textbf{T5}~\cite{raffel2020exploring}. 
    T5 is pre-trained on the Colossal Clean Crawled Corpus (C4), but no domain knowledge is specifically fed for pre-training.
    In this work, \emph{T5-base} and \emph{T5-large} are directly fine-tuned on ChEBI-20.
    
    \item \textbf{MolT5}~\cite{edwards-etal-2022-translation}. 
    MolT5 models are pre-trained on both language texts and SMILES strings.
    More specifically, \emph{MolT5-base} and \emph{MolT5-large} were pre-trained on the Colossal Clean Crawled Corpus (C4) and ZINC-15 datasets and further fine-tuned on ChEBI-20.

    
\end{itemize}

\subsection{Performance Comparison}
We present the results of the molecule-caption translation task, incorporating both quantitative analysis and detailed examples for comparison. 

\noindent\textbf{Molecule Captioning (Mol2Cap).}
Table \ref{tab:m2c} illustrates the performance comparison of 10-shot MolReGPT (GPT-3.5-turbo) and 10-shot MolReGPT (GPT-4-0314) with the baseline models for Mol2Cap task, offering an overview of the results. 
Notably, our method can achieve better BLEU scores and comparable ROUGE scores to MolT5-base when using GPT-3.5-turbo as the backend model. Meanwhile, when using GPT-4-0314 as the backend model, MolReGPT can obtain better BLEU and Text2Mol scores than MolT5-large without being fine-tuned on ChEBI-20 dataset. 
Furthermore, we obtain the following observations:
\begin{itemize} [leftmargin=*] 
\item With the instruction of 10-shot MolReGPT, GPT-3.5-turbo achieves significantly improved results that gain an improvement of 59\% to the zero-shot case and 2.4\% to MolT5-base under the Text2Mol metric, which indicates that our proposed method can teach ChatGPT to effectively learn the Mol2Cap task via in-context learning. 

\item Restricted by the number of examples, MolReGPT only gains limited insights from the distribution of molecule captions (e.g., vocabulary and grammar).
The LLMs' predictions for captions heavily rely on their internal factual knowledge and the contextual information provided by the system prompt. Thus, such vocabulary and grammar patterns may not be as apparent and can not be captured from the selected $n$ examples. 
As a result, although our 10-shot MolReGPT (GPT-4-0314) achieves a 0.593 Text2Mol score, which is higher than MolT5-large's 0.554. MolReGPT, in turn, gets lower ROUGE scores compared to MolT5.
However, it is crucial to note that captions generated by 10-shot MolReGPT with lower ROUGE scores are not entirely incorrect. 
In fact, the highest Text2Mol score serves as a reliable indicator of the generation quality and highlights the better relevance between the generated molecules and the molecule captions.
\end{itemize}

Considering the model agnosticism, MolReGPT increases the caption generation performance consistently across the three different models. This is particularly evident with Llama-2-7B, as it possesses a different model structure. As depicted in Table \ref{tab:m2c}, Llama-2-7B achieves 0.409 BLEU-4 score and 0.466 Text2Mol score without further fine-tuning, which is close to the performance of T5-base. This denotes that MolReGPT is not dependent on specific models but is a general method that could be applicable to a broad spectrum of LLMs.

\begin{figure*}[htbp]
   \centering
   \includegraphics[width=1.0\textwidth]{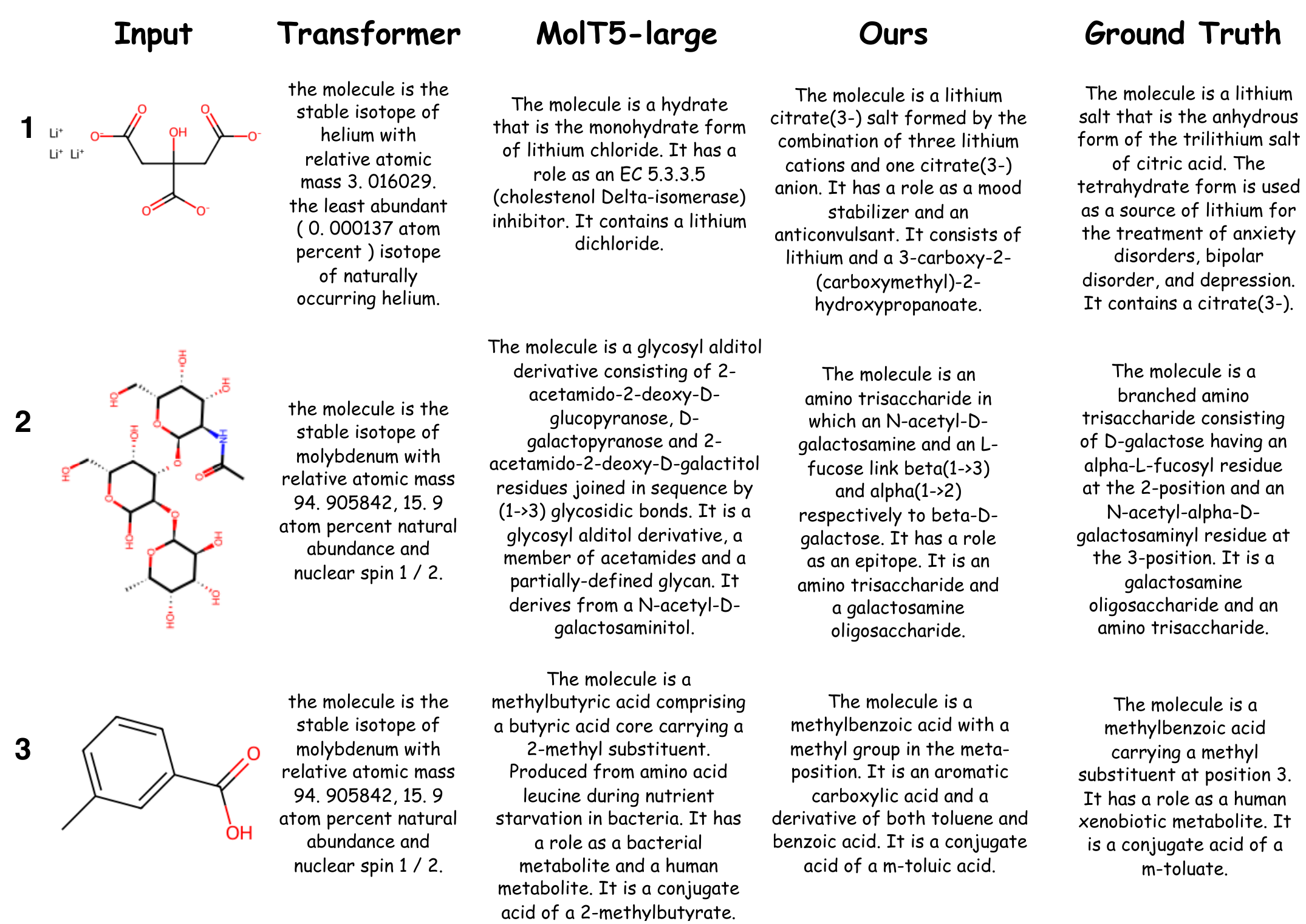} 
   \caption{Examples of molecule captions generated by different models, where SMILES strings are converted to molecule graphs for better visualization. 
   Based on the same input molecule graph, our MolReGPT (10-shot GPT-4-0314) can generate accurate and natural captions to describe the structure, properties, and even the functions of the molecule. 
   In contrast, Transformer generates meaningless captions that are far from the ground truth. Captions generated by MolT5-large seem better but still have some typo errors.}
   \label{fig:smiles2caption_diffmodels}
\end{figure*}
Figure \ref{fig:smiles2caption_diffmodels} compares our predicted captions with the ground truth. It is clear that our prediction is quite close to the ground truth, stating accurate details of molecule structures and giving solid predictions of molecule properties. However, it can also be seen that our results have some slight differences in the narrative order, which might influence the translation scores like BLEU and ROUGE. In Contrast, although MolT5-large achieves higher ROUGE scores, the generated captions still make many fact and typo errors, contributing to a lower Text2Mol score. Besides, Transformer could hardly generate correct or valid results. For molecule 2 and 3, it even generates the same results, which means that the model is not sensitive to the input.


\noindent\textbf{Text-based Molecule Generation (Cap2Mol).}
\begin{figure*}[htbp]
   \centering
   \includegraphics[width=1.0\textwidth]{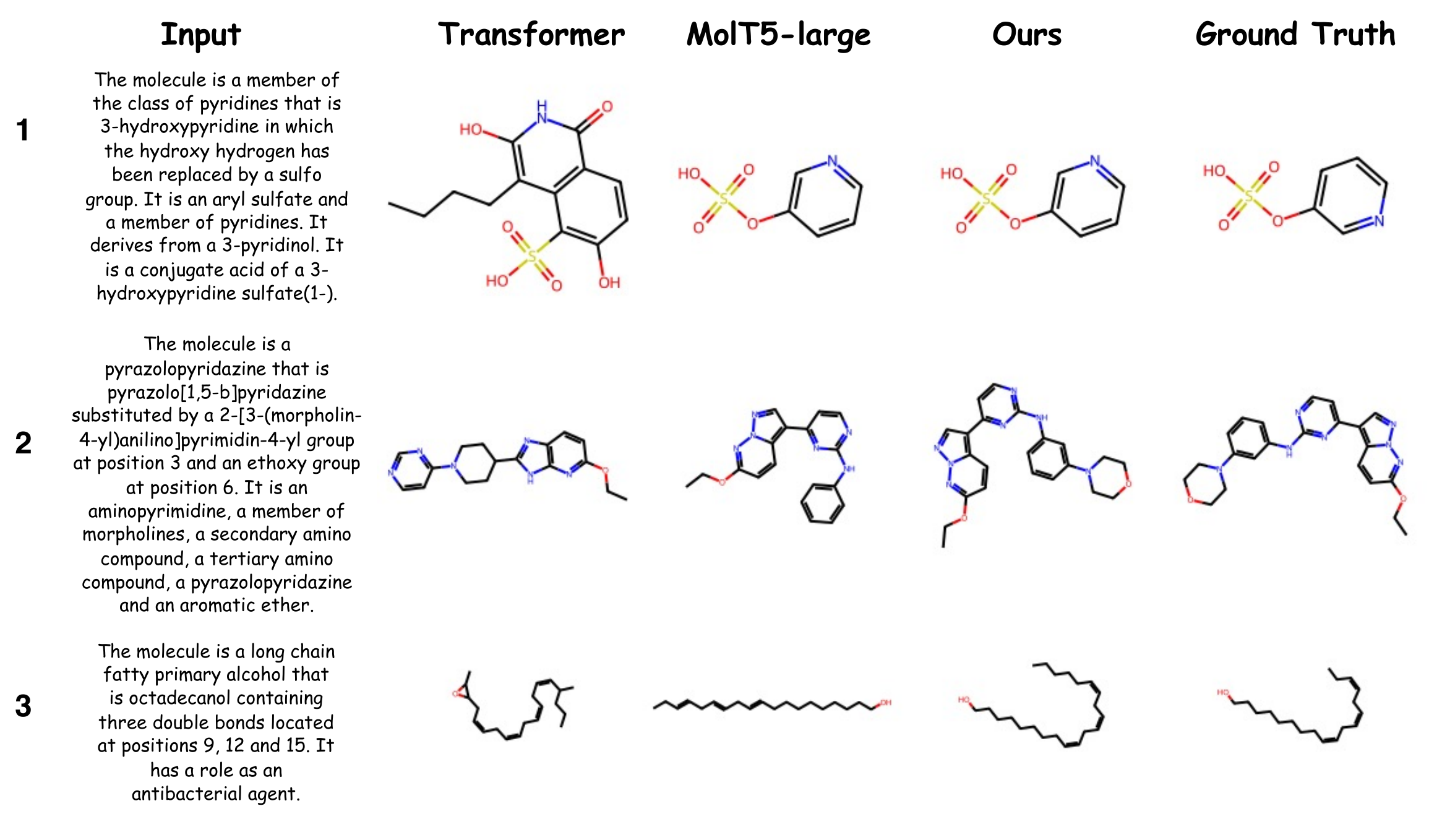}
   \caption{Examples of molecules generated by different models, where SMILES strings are converted to molecule graphs for better visualization. Based on the same input caption, our MolReGPT (10-shot GPT-4-0314) can generate accurate molecule graphs described by the caption. In contrast, Transformer generates quite different molecules compared to the ground truth. Compared to Transformer, molecules generated by MolT5-large are closer to the ground truth but still miss so many details.}
   \label{fig:caption2smiles_diffmodels}
\end{figure*}
Results of the text-based molecule generation task are presented in Table \ref{tab:c2m}. 
Comparing all these baselines, 10-shot MolReGPT significantly enhances the capabilities of GPT-3.5-turbo and GPT-4-0314, leading to improved overall performance.
In the Text2Mol metric, MolReGPT helps GPT-3.5-turbo and GPT-4-0314 gain a significant 15\% and 20\% increase, respectively, compared to MolT5-base. What's more, GPT-4-0314 even achieves a 7\% improvement compared to MolT5-large.
Considering the fingerprint scores, our 10-shot MolReGPT (GPT-4-0314) even gets an average of 8.1\% improvement compared to MolT5-large. 
More importantly, MolReGPT also significantly enhances the molecule generation capabilities of Llama-2-7B, elevating the BELU score from 0.104 to 0.693. Given that Llama-2-7B was trained on a distinct corpus and possesses a different model architecture, it could further demonstrate the model-agnostic nature of our approach.

Figure \ref{fig:caption2smiles_diffmodels} compares our predicted molecules with the ground truth. It can be seen that our generated molecules are quite close to the ground truth in the molecular configuration. For molecule 1, both MolT5-large and MolReGPT generate the exact correct molecule, but the 2D graph is slightly different to the ground truth due to the sequence order of the SMILES string. For molecule 2, MolT5-large misses several key structures, while MolReGPT generates the correct representation. For molecule 3, MolT5-large fails to generate the correct configuration, while MolReGPT generates the correct chemical bonds but misses the correct number of carbon atoms.

\begin{figure*}[htbp]
   \centering
   \includegraphics[width=1.0\textwidth]{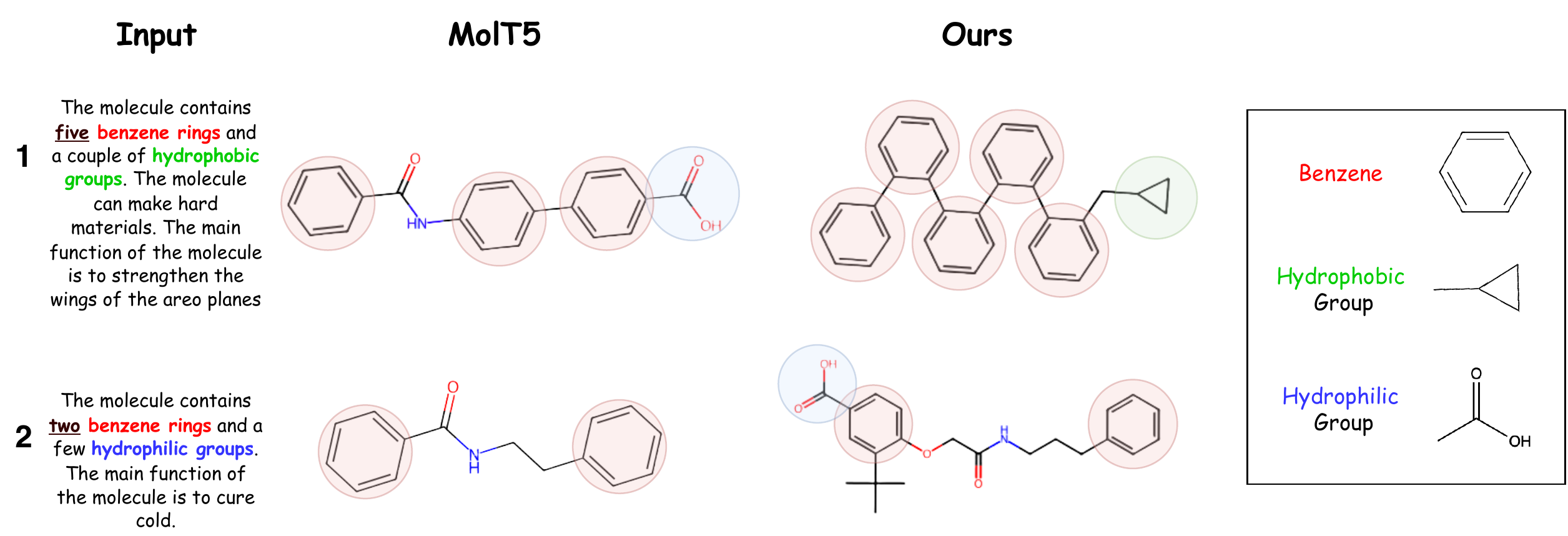}   
   \caption{
   Illustrations of molecule graphs generated by MolT5 and our MolReGPT (GPT-3.5-turbo), given  customized inputs.
   Notably, the key points in Example 1 highlight the \underline{\textbf{five}} \textbf{\red{benzene rings}} and \textbf{\green{hydrophobic groups}} in the structure, which are correctly generated by our MolReGPT. 
   In contrast, the results of MolT5 generate the incorrect number of \red{\textbf{benzene rings}} and contain a few \textbf{\blue{hydrophilic groups}}. 
   In example 2, both generations give the correct number of benzene rings, while MolReGPT generates more hydrophilic groups, which are closer to our input caption.}
   \label{fig:case study}
\end{figure*}

Besides, in Figure \ref{fig:case study}, we propose a practical scenario where a scientist aims to obtain molecules with desired structures and properties. In the past, scientist needs to apply domain knowledge to figure out a possible molecule candidate. Then, experiments are required to verify its properties. Now, with the help of MolReGPT, the scientist could formalize the requirements via molecule captions and ask MolReGPT to generate the desired molecules. 
To be more specific, we list two examples here. Molecule 1 has five benzene rings and several hydrophobic groups as its unique pattern, while molecule 2 has two benzene rings and several hydrophilic groups. By counting the number of benzene rings, we can easily find that MolT5 generates 3 benzene rings in molecule 1, which is incorrect. Besides, the functional groups generated by MolT5 in both molecule 1 and molecule 2 also miss-match the requirements given in the captions.
Remarkably, these impressive results are achieved without additional fine-tuning steps.
\begin{table*}[htbp]
    \centering

    \resizebox{2.0\columnwidth}{!}{
    \begin{tabular}{c|c|c|c|c|c|c|c}
    \toprule
    Method & BLEU-2$\uparrow$ & BLEU-4$\uparrow$ & ROUGE-1$\uparrow$ & ROUGE-2$\uparrow$ & ROUGE-L$\uparrow$ & METEOR$\uparrow$ & Text2Mol$\uparrow$ \\
    \midrule
    zero-shot & 0.103 & 0.050 & 0.261 & 0.088 & 0.204 & 0.161 & 0.352 \\ 
    \hline
    1-shot (random) & 0.236 & 0.131 & 0.335 & 0.135 & 0.257 & 0.253 & 0.372 \\ 
    1-shot (BM25) & 0.243 & 0.150 & 0.350 & 0.156 & 0.278 & 0.262 & 0.394\\ 
    1-shot (Morgan FTS) & 0.506 & 0.416 & 0.547 & 0.372 & 0.473 & 0.499 & 0.529\\ 
    \hline
    2-shot (random) & 0.273 & 0.158 & 0.357 & 0.154 & 0.278 & 0.284 & 0.371 \\ 
    2-shot (BM25) & 0.287 & 0.188 & 0.380 & 0.185 & 0.307 & 0.297 & 0.397 \\ 
    2-shot (Morgan FTS) & 0.547 & 0.460 & 0.592 & 0.425 & 0.520 & 0.559 & 0.548 \\ 
    \hline
    5-shot (random) & 0.297 & 0.178 & 0.376 & 0.173 & 0.300 & 0.305 & 0.366 \\ 
    5-shot (BM25) & 0.311 & 0.213 & 0.398 & 0.205 & 0.327 & 0.317 & 0.405 \\ 
    5-shot (Morgan FTS) & \underline{0.562} & \underline{0.478} & \underline{0.609} & \underline{0.446} & \underline{0.540} & \underline{0.583} & \underline{0.559(6)} \\ 
    \hline
    10-shot (random) & 0.295 & 0.181 & 0.389 & 0.185 & 0.310 & 0.329 & 0.369 \\ 
    10-shot (BM25) & 0.326 & 0.227 & 0.413 & 0.221 & 0.342 & 0.333 & 0.408 \\ 
    10-shot (Morgan FTS) & \textbf{0.565} & \textbf{0.482} & \textbf{0.623} & \textbf{0.450} & \textbf{0.543} & \textbf{0.585} & \textbf{0.559(8)} \\ 
    \bottomrule
    \end{tabular}
    }

    \caption{N-shot Molecule Captioning results on ChEBI-20 dataset with the backend model, GPT-3.5-turbo. The \textbf{best} scores are in bold, and the \underline{second-best} scores are underlined.}

    \label{tab:m2c_all}
\end{table*}

\begin{table*}[htbp]
    \centering
    \resizebox{2.0\columnwidth}{!}{
    \begin{tabular}{c|c|c|c|c|c|c|c|c|c}
    \toprule
    Method & BLEU$\uparrow$ & EM$\uparrow$ & Levenshtein$\downarrow$ & MACCS FTS$\uparrow$ & RDK FTS$\uparrow$ & Morgan FTS$\uparrow$ & FCD$\downarrow$ & Text2Mol$\uparrow$ & Validity$\uparrow$ \\
    \midrule
    zero-shot & 0.489 & 0.019 & 52.13 & 0.705 & 0.462 & 0.367 & 2.05 & 0.479 & 0.802 \\ 
    \hline
    1-shot (random) & 0.525 & 0.027 & 51.86 & 0.716 & 0.475 & 0.373 & 1.67 & 0.482 & 0.821\\ 
    1-shot (SentenceBert) & 0.687 & 0.066 & 35.89 & 0.796 & 0.609 & 0.511 & 0.85 & 0.541 & 0.839\\ 
    1-shot (BM25) & 0.706 & 0.074 & 33.38 & 0.799 & 0.620 & 0.526 & 0.84 & 0.540 & 0.842\\ 
    \hline
    2-shot (random) & 0.529 & 0.026 & 49.87 & 0.720 & 0.479 & 0.380 & 1.71 & 0.483 & 0.824\\ 
    2-shot (SentenceBert) & 0.642 & 0.048 & 40.98 & 0.770 & 0.560 & 0.463 & 1.01 & 0.557 & 0.841\\ 
    2-shot (BM25) & 0.748 & 0.101 & 28.89 & 0.827 & 0.668 & 0.578 & 0.67 & 0.519 & 0.860\\ 
    \hline
    5-shot (random) & 0.552 & 0.028 & 49.26 & 0.720 & 0.476 & 0.382 & 1.60 & 0.481 & 0.832\\ 
    5-shot (SentenceBert) & 0.758 & 0.095 & 28.34 & 0.824 & 0.659 & 0.568 & 0.71 & 0.558 & 0.871\\ 
    5-shot (BM25) & \underline{0.771} & \underline{0.121} & \underline{26.78} & \underline{0.836} & \underline{0.686} & \underline{0.599} & \underline{0.60} & \underline{0.564} & 0.882\\ 
    \hline
    10-shot (random) & 0.564 & 0.029 & 49.11 & 0.723 & 0.486 & 0.386 & 1.46 & 0.484 & 0.846\\ 
    10-shot (SentenceBert) & 0.767 & 0.098 & 27.46 & 0.831 & 0.672 & 0.585 & 0.63 & 0.562 & \textbf{0.890}\\ 
    10-shot (BM25) & \textbf{0.790} & \textbf{0.139} & \textbf{24.91} & \textbf{0.847} & \textbf{0.708} & \textbf{0.624} & \textbf{0.57} & \textbf{0.571} & \underline{0.887} \\ 
    \bottomrule
    \end{tabular}
    }

    \caption{N-shot Molecule Generation results on ChEBI-20 dataset with the backend model, GPT-3.5-turbo. The \textbf{best} scores are in bold, and the \underline{second-best} scores are \underline{underlined}.}
    \label{tab:c2m_all}

\end{table*}
Furthermore, the original weights of T5 are primarily for natural language, which means it has to be fine-tuned separately to fit the two sub-tasks in this study. 
Unfortunately, MolT5 does not tackle this issue, as it continues to treat the two sub-tasks of the molecule-caption translation task separately.
Switching between the two sub-tasks in MolT5 requires using a different model class and reloading the weights, making it technically inefficient. 
In contrast, MolReGPT enables a single foundation LLM to solve both the two sub-tasks simultaneously, providing a comprehensive solution for LLMs to address molecule-related tasks.


\subsection{Ablation Study}
In addition to the experiments above, we also perform ablation studies to analyze the critical factors that influence the performance of MolReGPT. Note that we select GPT-3.5-turbo as the backend model. 

\subsubsection{Impact of Retrieval Strategies.}
Retrieval strategies play a key role in guiding LLMs to perform molecule-caption translation tasks for MolReGPT. 
More similar examples are retrieved, and more valuable information could be contained for In-Context Few-Shot Molecule Learning.
For Mol2Cap and Cap2Mol, we choose three different retrieval strategies for comparison. The detailed results are shown in Table \ref{tab:m2c_all} and Table \ref{tab:c2m_all}.  We show that in both sub-tasks, compared to random selection, the other retrieval strategies used in this paper can help improve n-shot generation results. Thus, thoughtful selection of retrieval strategies plays a key role in MolReGPT.

In \textbf{Mol2Cap} task, 
we compare the performance of three retrieval strategies: Random, BM25, and Morgan FTS (adopted in MolReGPT). 
The Random strategy involves retrieving \(n\) random examples, while BM25 applies a character-level BM25 algorithm to the molecule SMILES representations.
As shown in Table \ref{tab:m2c_all}, among the three strategies, Morgan FTS shows the best performance when $n$ is fixed, outperforming BM25 by 37\% in the Text2Mol metric. Besides, the ROUGE-L score achieved by Morgan FTS is almost doubled compared to the Random or BM25 strategies. 
The use of Morgan FTS with Dice similarity shows a better estimation of the structural similarity between molecules by comparing unique structural features like functional groups. These features are usually revealed in molecule captions with detailed descriptions. In this case, retrieving similar molecules by Morgan FTS could effectively guide the LLM to learn the associations between molecule structures and caption descriptions, resulting in more accurate and desired outputs.

In \textbf{Cap2Mol} task, we also employ these retrieval strategies: Random, SentenceBert, and BM25 (adopted in MolReGPT). The Random strategy still retrieves \(n\) random examples, while SentenceBert encodes captions as numerical vectors to compute their semantic similarity.
As shown in Table \ref{tab:c2m_all}, BM25 is the best one in the Cap2Mol task, despite the fact that SentenceBert has outperformed BM25 in many classical NLP text retrieval datasets. 
When $n$ changes from 1 to 10, BM25 always achieves better BLEU, Exact Match, Levenshtein, and fingerprint scores than SentenceBert.
The input molecule captions tend to use phrases with dashes (-) like "2-methylphenyl" to connect the structure details of the molecule. 
Understanding such phrases plays a crucial role in generating correct molecule structures. 
In this case, retrieving similar texts while precisely matching these details significantly contributes to performance improvement. In contrast, SentenceBert, as a neural method, encodes an entire caption into a 1-D embedding vector, focusing more on semantic similarity rather than specific details.
Consequently, BM25 is better than SentenceBert in the Cap2Mol task.


\subsubsection{Impact of Example Number for In-Context Learning.}
In this subsection, we study how the number of examples contained in the system prompt through in-context learning affects the performance.

In the \textbf{zero-shot scenario}, where no extra examples are included in the prompt for guiding LLMs for learning the molecule-caption translation task, we utilize two special spans, `[MOLECULE\_MASK]' and `[CAPTION\_MASK]', to inform the LLMs of the desired output format. 

After analyzing the zero-shot results of GPT-3.5-turbo in Tables \ref{tab:m2c_all} and \ref{tab:c2m_all}, we can observe that SMILES strings are included in its pre-training corpus because it can generate basically valid SMILES representations of molecules based on zero-shot prompts, achieving a 0.802 validity score and a 0.479 Text2Mol score in molecule generation.
However, it is important to note that the zero-shot results exhibit a performance level similar to a vanilla Transformer model. This observation provides evidence that GPT-3.5-turbo is not specifically trained on ChEBI-20, thereby alleviating concerns regarding potential information leakage.

For \textbf{Few-shot Performance},
Table \ref{tab:m2c_all} and Table \ref{tab:c2m_all} list the comprehensive details of the experimental results.
Normally, the performance improves as the number of examples, denoted as \(n\), increases, as more examples provide additional knowledge for the task. 
However, due to the input length limitation of LLMs, it is impossible to contain a large number of examples in the system prompt. 
Therefore, for few-shot scenarios, we choose four different values 1, 2, 5, and 10. 

Tables \ref{tab:m2c_all} and \ref{tab:c2m_all} illustrate that performance generally improves as $n$ increases through in-context learning. Significant performance enhancements are observed as n changes from 0 to 10. 
Taking Morgan FTS and BM25 as examples, in caption generation, we see remarkable increases from 0.050 to 0.482, 0.204 to 0.543, and 0.352 to 0.560 in BLEU-4, ROUGE-L, and Text2Mol scores, respectively.
Besides, BM25 improves molecule generation from 0.489 to 0.790 in the BLEU score and 0.479 to 0.571 in the Text2Mol score.

Besides, it is worth noticing that when $n$ increases from 5 to 10, the Text2Mol metrics almost keep the same, which can be the problem of the maximum input length limitation of LLMs. To fit the input length limitation, we would remove the longest examples to degrade the n-shot to (n-1)-shot generation for re-queries. As $n$ increases, there is a higher possibility of exceeding the input length limitation. In this case, unless the maximum input length of the LLM is expanded, the performance will finally converge when $n$ continues to grow.

\section{Conclusion}
\label{sec:conclusion}
In this work, we propose MolReGPT, a general retrieval-based in-context learning paradigm that empowers molecule discovery with LLMs like ChatGPT via In-Context Few-Shot Molecule Learning. Our method is focused and evaluated on the task of molecule-caption translation, including molecule captioning (Mol2Cap) and text-based molecule generation (Cap2Mol). MolReGPT leverages the molecular similarity principle to retrieve examples from a local database, guiding LLMs to generate predictions without being fine-tuned. 
Specifically, BM25 caption retrieval is applied to obtain similar molecule captions, while Morgan Fingerprints and Dice similarity are adopted to retrieve similar molecules. 
Experimental results show that our proposed MolReGPT can empower ChatGPT to achieve 0.585 and 0.593 Text2Mol scores in Mol2Cap and Cap2Mol, respectively.
Compared to MolT5-large, our MolReGPT equips GPT-4-0314 with the ability to achieve comparable performance in Mol2Cap task and even outperform MolT5-large in Cap2Mol task without any fine-tuning steps.
To conclude, MolReGPT provides a novel and versatile paradigm to deploy LLMs in molecule discovery through in-context learning, which greatly reduces the cost of domain transfer and explores the potential of LLMs in molecule discovery.

\section{Broader Implication \& Future Directions}
Our work has demonstrated the effectiveness of the MolReGPT framework across various LLMs. Compared to the existing methods, MolReGPT does not require additional pre-training or fine-tuning, yet it enables powerful LLMs like GPT-4 to achieve comparable and even superior performance. The integration of LLMs and biomolecular science represents a paradigm shift in molecule discovery. As illustrated in Figure \ref{fig:case study}, fine-tuned methods tend to fail to meet the requirements in customized inputs, while MolReGPT shows a better generalization potential. 
These unique features make our method more convenient and practical for chemists to adopt and use in their work. 
We hope that MolReGPT could enable a wider range of scientific researchers to leverage the power of LLMs in their work, ultimately contributing to the advancement of scientific research and discovery.

For future work, there are still several areas that require further exploration and improvement.
\emph{Firstly,} the performance of LLMs is intricately related to the prediction quality of MolReGPT. Thus, with the advancement of more powerful LLMs, our methods could potentially yield even better results in the future.
\emph{Secondly,} we could develop better retrieval algorithms that could help refine the context examples. For example, we could combine BM25 caption retrieval with chemical LLMs and apply Graph Neural Networks for molecular similarity to improve the retrieval quality.
\emph{Lastly,} we anticipate that our work will not only make LLMs more accessible for scientific researchers in the field of chemistry, thereby benefiting drug discovery, but also inspire the AI community to consider the alignment between molecular and text space.

\bibliographystyle{IEEEtran}

\bibliography{reference}

\begin{thebibliography}{10}
\providecommand{\url}[1]{#1}
\csname url@samestyle\endcsname
\providecommand{\newblock}{\relax}
\providecommand{\bibinfo}[2]{#2}
\providecommand{\BIBentrySTDinterwordspacing}{\spaceskip=0pt\relax}
\providecommand{\BIBentryALTinterwordstretchfactor}{4}
\providecommand{\BIBentryALTinterwordspacing}{\spaceskip=\fontdimen2\font plus
\BIBentryALTinterwordstretchfactor\fontdimen3\font minus \fontdimen4\font\relax}
\providecommand{\BIBforeignlanguage}[2]{{%
\expandafter\ifx\csname l@#1\endcsname\relax
\typeout{** WARNING: IEEEtran.bst: No hyphenation pattern has been}%
\typeout{** loaded for the language `#1'. Using the pattern for}%
\typeout{** the default language instead.}%
\else
\language=\csname l@#1\endcsname
\fi
#2}}
\providecommand{\BIBdecl}{\relax}
\BIBdecl

\bibitem{xu2023plasma}
J.~Xu, Y.~Li, J.~Yang, S.~Zhou, and W.~Situ, ``Plasma etching effect on the molecular structure of chitosan-based hydrogels and its biological properties,'' \emph{International Journal of Biological Macromolecules}, p. 123257, 2023.

\bibitem{weng2021late}
Y.~Weng, B.~Ding, Y.~Liu, C.~Song, L.-Y. Chan, and C.-W. Chiang, ``Late-stage photoredox c--h amidation of n-unprotected indole derivatives: Access to n-(indol-2-yl) amides,'' \emph{Organic Letters}, vol.~23, no.~7, pp. 2710--2714, 2021.

\bibitem{ding2019selective}
B.~Ding, Y.~Weng, Y.~Liu, C.~Song, L.~Yin, J.~Yuan, Y.~Ren, A.~Lei, and C.-W. Chiang, ``Selective photoredox trifluoromethylation of tryptophan-containing peptides,'' \emph{European Journal of Organic Chemistry}, vol. 2019, no.~46, pp. 7596--7605, 2019.

\bibitem{higuchi2023material}
A.~Higuchi, T.-C. Sung, T.~Wang, Q.-D. Ling, S.~S. Kumar, S.-T. Hsu, and A.~Umezawa, ``Material design for next-generation mrna vaccines using lipid nanoparticles,'' \emph{Polymer Reviews}, vol.~63, no.~2, pp. 394--436, 2023.

\bibitem{urbina2022commoditization}
F.~Urbina and S.~Ekins, ``The commoditization of ai for molecule design,'' \emph{Artificial Intelligence in the Life Sciences}, vol.~2, p. 100031, 2022.

\bibitem{weininger1988smiles}
D.~Weininger, ``Smiles, a chemical language and information system. 1. introduction to methodology and encoding rules,'' \emph{Journal of chemical information and computer sciences}, vol.~28, no.~1, pp. 31--36, 1988.

\bibitem{arus2019randomized}
J.~Ar{\'u}s-Pous, S.~V. Johansson, O.~Prykhodko, E.~J. Bjerrum, C.~Tyrchan, J.-L. Reymond, H.~Chen, and O.~Engkvist, ``Randomized smiles strings improve the quality of molecular generative models,'' \emph{Journal of cheminformatics}, vol.~11, no.~1, pp. 1--13, 2019.

\bibitem{honda2019smiles}
S.~Honda, S.~Shi, and H.~R. Ueda, ``Smiles transformer: Pre-trained molecular fingerprint for low data drug discovery,'' \emph{arXiv preprint arXiv:1911.04738}, 2019.

\bibitem{edwards2021text2mol}
C.~Edwards, C.~Zhai, and H.~Ji, ``Text2mol: Cross-modal molecule retrieval with natural language queries,'' in \emph{Proceedings of the 2021 Conference on Empirical Methods in Natural Language Processing}, 2021, pp. 595--607.

\bibitem{edwards-etal-2022-translation}
\BIBentryALTinterwordspacing
C.~Edwards, T.~Lai, K.~Ros, G.~Honke, K.~Cho, and H.~Ji, ``Translation between molecules and natural language,'' in \emph{Proceedings of the 2022 Conference on Empirical Methods in Natural Language Processing}.\hskip 1em plus 0.5em minus 0.4em\relax Abu Dhabi, United Arab Emirates: Association for Computational Linguistics, Dec. 2022, pp. 375--413. [Online]. Available: \url{https://aclanthology.org/2022.emnlp-main.26}
\BIBentrySTDinterwordspacing

\bibitem{su2022molecular}
B.~Su, D.~Du, Z.~Yang, Y.~Zhou, J.~Li, A.~Rao, H.~Sun, Z.~Lu, and J.-R. Wen, ``A molecular multimodal foundation model associating molecule graphs with natural language,'' \emph{arXiv preprint arXiv:2209.05481}, 2022.

\bibitem{zhu2023minigpt}
D.~Zhu, J.~Chen, X.~Shen, X.~Li, and M.~Elhoseiny, ``Minigpt-4: Enhancing vision-language understanding with advanced large language models,'' \emph{arXiv preprint arXiv:2304.10592}, 2023.

\bibitem{bao2023tallrec}
K.~Bao, J.~Zhang, Y.~Zhang, W.~Wang, F.~Feng, and X.~He, ``Tallrec: An effective and efficient tuning framework to align large language model with recommendation,'' \emph{arXiv preprint arXiv:2305.00447}, 2023.

\bibitem{rubin2022learning}
O.~Rubin, J.~Herzig, and J.~Berant, ``Learning to retrieve prompts for in-context learning,'' in \emph{Proceedings of the 2022 Conference of the North American Chapter of the Association for Computational Linguistics: Human Language Technologies}, 2022, pp. 2655--2671.

\bibitem{min2022metaicl}
S.~Min, M.~Lewis, L.~Zettlemoyer, and H.~Hajishirzi, ``Metaicl: Learning to learn in context,'' in \emph{Proceedings of the 2022 Conference of the North American Chapter of the Association for Computational Linguistics: Human Language Technologies}, 2022, pp. 2791--2809.

\bibitem{hu2023deep}
W.~Hu, Y.~Liu, X.~Chen, W.~Chai, H.~Chen, H.~Wang, and G.~Wang, ``Deep learning methods for small molecule drug discovery: A survey,'' \emph{IEEE Transactions on Artificial Intelligence}, 2023.

\bibitem{fan2023generative}
W.~Fan, C.~Liu, Y.~Liu, J.~Li, H.~Li, H.~Liu, J.~Tang, and Q.~Li, ``Generative diffusion models on graphs: Methods and applications,'' \emph{arXiv preprint arXiv:2302.02591}, 2023.

\bibitem{peng2019convolutional}
S.-P. Peng and Y.~Zhao, ``Convolutional neural networks for the design and analysis of non-fullerene acceptors,'' \emph{Journal of Chemical Information and Modeling}, vol.~59, no.~12, pp. 4993--5001, 2019.

\bibitem{le2019imotor}
N.~Q.~K. Le, E.~K.~Y. Yapp, Y.-Y. Ou, and H.-Y. Yeh, ``imotor-cnn: Identifying molecular functions of cytoskeleton motor proteins using 2d convolutional neural network via chou's 5-step rule,'' \emph{Analytical biochemistry}, vol. 575, pp. 17--26, 2019.

\bibitem{grisoni2020bidirectional}
F.~Grisoni, M.~Moret, R.~Lingwood, and G.~Schneider, ``Bidirectional molecule generation with recurrent neural networks,'' \emph{Journal of chemical information and modeling}, vol.~60, no.~3, pp. 1175--1183, 2020.

\bibitem{amabilino2020guidelines}
S.~Amabilino, P.~Pog{\'a}ny, S.~D. Pickett, and D.~V. Green, ``Guidelines for recurrent neural network transfer learning-based molecular generation of focused libraries,'' \emph{Journal of Chemical Information and Modeling}, vol.~60, no.~12, pp. 5699--5713, 2020.

\bibitem{bagal2021molgpt}
V.~Bagal, R.~Aggarwal, P.~Vinod, and U.~D. Priyakumar, ``Molgpt: molecular generation using a transformer-decoder model,'' \emph{Journal of Chemical Information and Modeling}, vol.~62, no.~9, pp. 2064--2076, 2021.

\bibitem{wang2021multi}
J.~Wang, C.-Y. Hsieh, M.~Wang, X.~Wang, Z.~Wu, D.~Jiang, B.~Liao, X.~Zhang, B.~Yang, Q.~He \emph{et~al.}, ``Multi-constraint molecular generation based on conditional transformer, knowledge distillation and reinforcement learning,'' \emph{Nature Machine Intelligence}, vol.~3, no.~10, pp. 914--922, 2021.

\bibitem{zeng2022deep}
Z.~Zeng, Y.~Yao, Z.~Liu, and M.~Sun, ``A deep-learning system bridging molecule structure and biomedical text with comprehension comparable to human professionals,'' \emph{Nature communications}, vol.~13, no.~1, p. 862, 2022.

\bibitem{radford2018improving}
A.~Radford, K.~Narasimhan, T.~Salimans, I.~Sutskever \emph{et~al.}, ``Improving language understanding by generative pre-training,'' \emph{OpenAI}, 2018.

\bibitem{radford2019language}
A.~Radford, J.~Wu, R.~Child, D.~Luan, D.~Amodei, I.~Sutskever \emph{et~al.}, ``Language models are unsupervised multitask learners,'' \emph{OpenAI blog}, vol.~1, no.~8, p.~9, 2019.

\bibitem{ouyang2022training}
L.~Ouyang, J.~Wu, X.~Jiang, D.~Almeida, C.~Wainwright, P.~Mishkin, C.~Zhang, S.~Agarwal, K.~Slama, A.~Ray \emph{et~al.}, ``Training language models to follow instructions with human feedback,'' \emph{Advances in Neural Information Processing Systems}, vol.~35, pp. 27\,730--27\,744, 2022.

\bibitem{openai-chatgpt}
OpenAI, ``Introducing chatgpt,'' 2022, \url{https://openai.com/blog/chatgpt}.

\bibitem{thoppilan2022lamda}
R.~Thoppilan, D.~De~Freitas, J.~Hall, N.~Shazeer, A.~Kulshreshtha, H.-T. Cheng, A.~Jin, T.~Bos, L.~Baker, Y.~Du \emph{et~al.}, ``Lamda: Language models for dialog applications,'' \emph{arXiv preprint arXiv:2201.08239}, 2022.

\bibitem{chowdhery2022palm}
A.~Chowdhery, S.~Narang, J.~Devlin, M.~Bosma, G.~Mishra, A.~Roberts, P.~Barham, H.~W. Chung, C.~Sutton, S.~Gehrmann \emph{et~al.}, ``Palm: Scaling language modeling with pathways,'' \emph{arXiv preprint arXiv:2204.02311}, 2022.

\bibitem{chiang2023vicuna}
W.-L. Chiang, Z.~Li, Z.~Lin, Y.~Sheng, Z.~Wu, H.~Zhang, L.~Zheng, S.~Zhuang, Y.~Zhuang, J.~E. Gonzalez \emph{et~al.}, ``Vicuna: An open-source chatbot impressing gpt-4 with 90\%* chatgpt quality,'' \emph{See https://vicuna. lmsys. org (accessed 14 April 2023)}, 2023.

\bibitem{bran2023chemcrow}
A.~M. Bran, S.~Cox, A.~D. White, and P.~Schwaller, ``Chemcrow: Augmenting large-language models with chemistry tools,'' \emph{arXiv preprint arXiv:2304.05376}, 2023.

\bibitem{white2023future}
A.~D. White, ``The future of chemistry is language,'' \emph{Nature Reviews Chemistry}, pp. 1--2, 2023.

\bibitem{chithrananda2020chemberta}
S.~Chithrananda, G.~Grand, and B.~Ramsundar, ``Chemberta: Large-scale self-supervised pretraining for molecular property prediction,'' \emph{arXiv preprint arXiv:2010.09885}, 2020.

\bibitem{liu2022multi}
S.~Liu, W.~Nie, C.~Wang, J.~Lu, Z.~Qiao, L.~Liu, J.~Tang, C.~Xiao, and A.~Anandkumar, ``Multi-modal molecule structure-text model for text-based retrieval and editing,'' \emph{arXiv preprint arXiv:2212.10789}, 2022.

\bibitem{frey2022neural}
N.~Frey, R.~Soklaski, S.~Axelrod, S.~Samsi, R.~Gomez-Bombarelli, C.~Coley, and V.~Gadepally, ``Neural scaling of deep chemical models,'' \emph{chemrxiv}, 2022.

\bibitem{wang2016improving}
Z.~Wang, L.~Liang, Z.~Yin, and J.~Lin, ``Improving chemical similarity ensemble approach in target prediction,'' \emph{Journal of cheminformatics}, vol.~8, pp. 1--10, 2016.

\bibitem{butina1999unsupervised}
D.~Butina, ``Unsupervised data base clustering based on daylight's fingerprint and tanimoto similarity: A fast and automated way to cluster small and large data sets,'' \emph{Journal of Chemical Information and Computer Sciences}, vol.~39, no.~4, pp. 747--750, 1999.

\bibitem{robertson2009probabilistic}
S.~Robertson, H.~Zaragoza \emph{et~al.}, ``The probabilistic relevance framework: Bm25 and beyond,'' \emph{Foundations and Trends{\textregistered} in Information Retrieval}, vol.~3, no.~4, pp. 333--389, 2009.

\bibitem{dice1945measures}
L.~R. Dice, ``Measures of the amount of ecologic association between species,'' \emph{Ecology}, vol.~26, no.~3, pp. 297--302, 1945.

\bibitem{coupry2022application}
D.~E. Coupry and P.~Pog{\'a}ny, ``Application of deep metric learning to molecular graph similarity,'' \emph{Journal of Cheminformatics}, vol.~14, no.~1, pp. 1--12, 2022.

\bibitem{aizawa2003information}
A.~Aizawa, ``An information-theoretic perspective of tf--idf measures,'' \emph{Information Processing \& Management}, vol.~39, no.~1, pp. 45--65, 2003.

\bibitem{touvron2023llama}
H.~Touvron, T.~Lavril, G.~Izacard, X.~Martinet, M.-A. Lachaux, T.~Lacroix, B.~Rozi{\`e}re, N.~Goyal, E.~Hambro, F.~Azhar \emph{et~al.}, ``Llama: Open and efficient foundation language models,'' \emph{arXiv preprint arXiv:2302.13971}, 2023.

\bibitem{vaswani2017attention}
A.~Vaswani, N.~Shazeer, N.~Parmar, J.~Uszkoreit, L.~Jones, A.~N. Gomez, {\L}.~Kaiser, and I.~Polosukhin, ``Attention is all you need,'' \emph{Advances in neural information processing systems}, vol.~30, 2017.

\bibitem{raffel2020exploring}
C.~Raffel, N.~Shazeer, A.~Roberts, K.~Lee, S.~Narang, M.~Matena, Y.~Zhou, W.~Li, and P.~J. Liu, ``Exploring the limits of transfer learning with a unified text-to-text transformer,'' \emph{The Journal of Machine Learning Research}, vol.~21, no.~1, pp. 5485--5551, 2020.

\end{thebibliography}

{\begin{IEEEbiography}[{\includegraphics[width=1in,height=1.25in,clip,keepaspectratio]{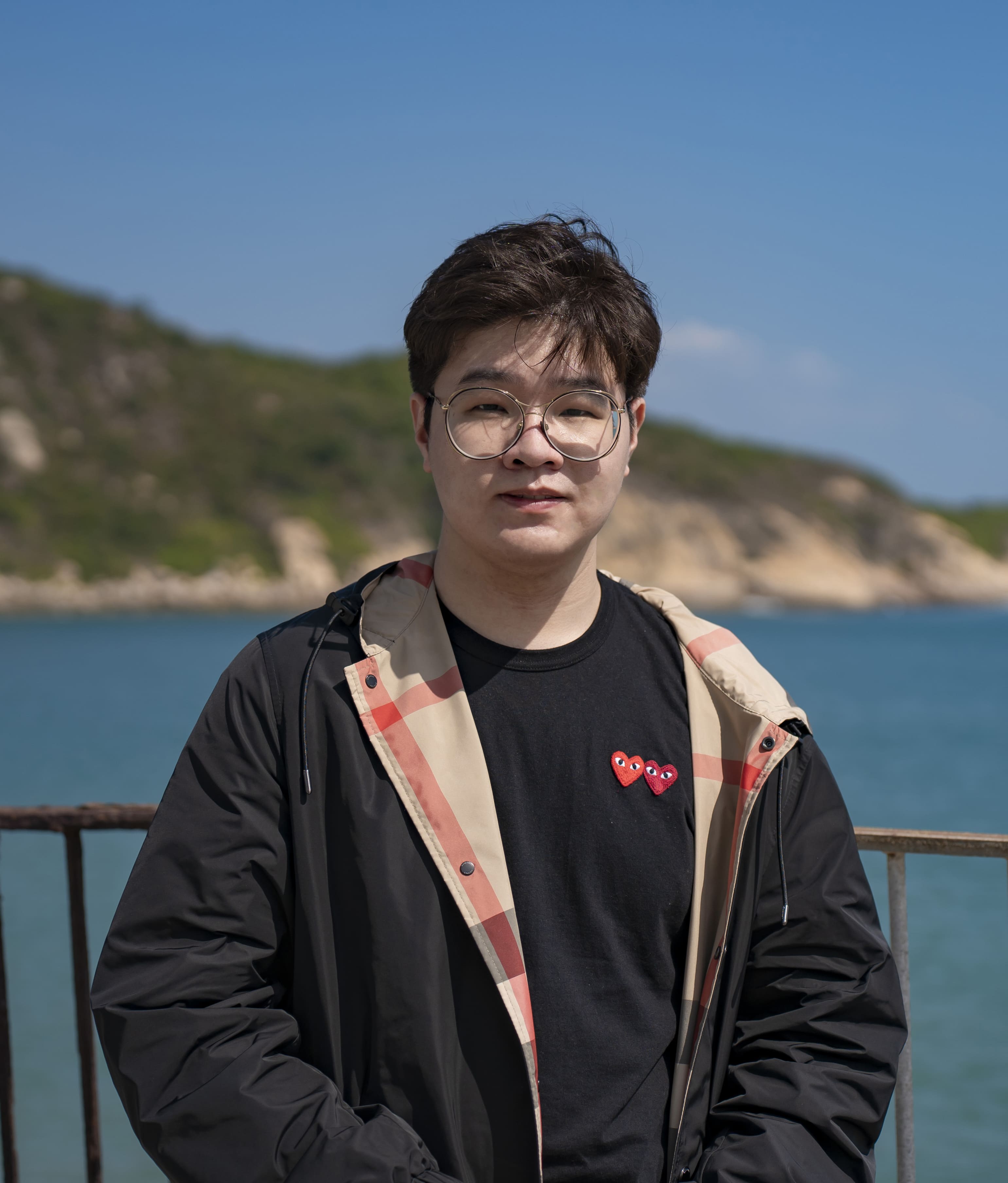}}]{Jiatong Li} is currently a PhD student of the Department of Computing (COMP), The Hong Kong Polytechnic University (funded by HKPFS). Before joining the PolyU, he received my Master's degree of Information Technology (with Distinction) from the University of Melbourne, under the supervision of Dr. Lea Frermann. In 2021, he got his bachelor's degree in Information Security from Shanghai Jiao Tong University. His interest lies in Natural Language Processing, Drug Discovery, and Recommender Systems. He has published innovative works in top-tier conferences such as IJCAI and ACL. For more information, please visit https://phenixace.github.io/.

\end{IEEEbiography}

\vspace{-24pt}
{\begin{IEEEbiography}[{\includegraphics[width=1in,height=1.25in,clip,keepaspectratio]{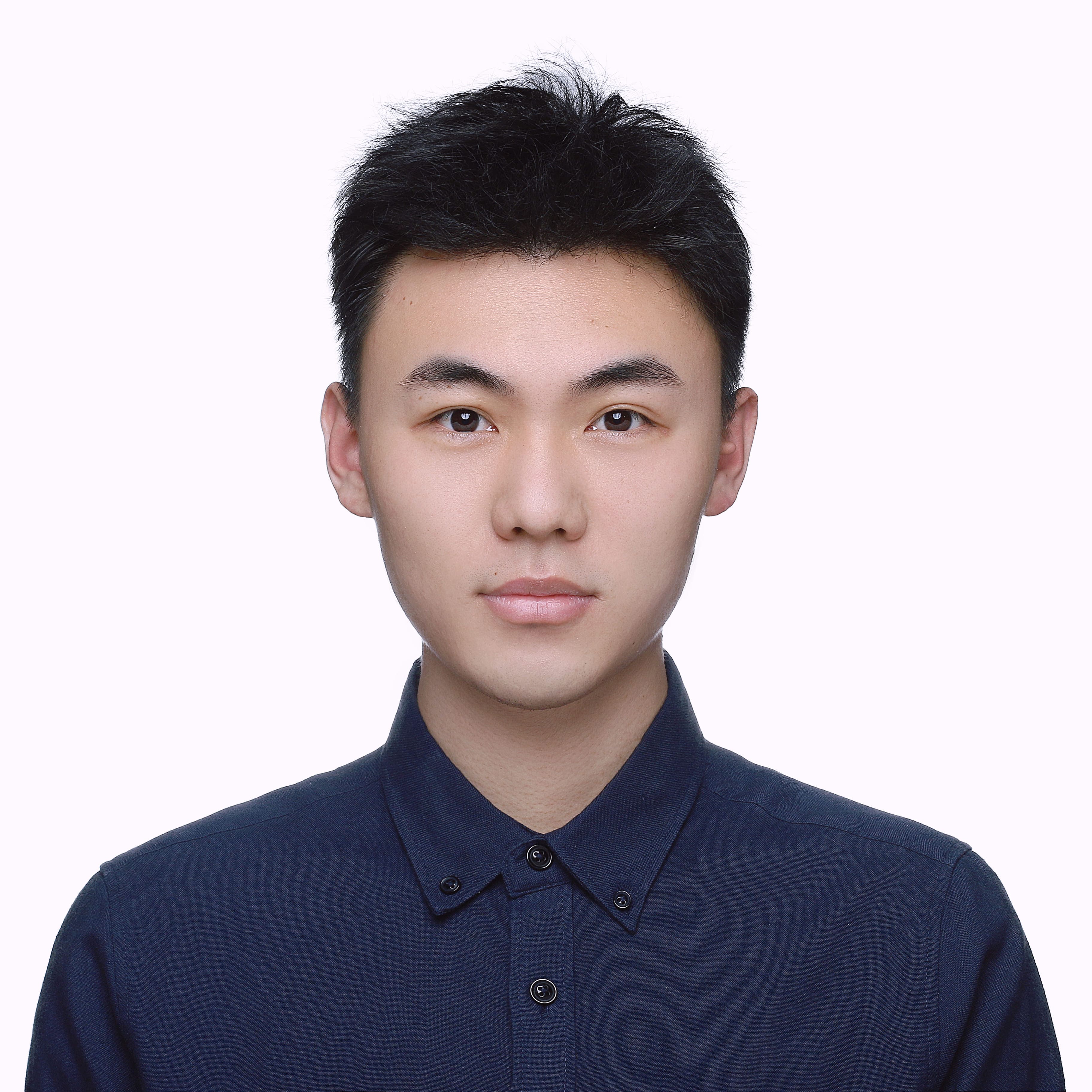}}]{Yunqing Liu} is currently a PhD student of the Department of Computing (COMP), Hong Kong Polytechnic University (PolyU), under the supervision of Dr. Wenqi Fan. Before joining the PolyU, he received his Master’s degree in Computer Science from the University of Edinburgh (M.Sc. in Computer Science), under the supervision of Dr. Elizabeth Polgreen. In 2020, he got his bachelor’s degrees from Wuhan University (B.Sc. in Chemistry and B.Eng. in Computer Science and Technology). His research interest includes Drug Discovery, Graph Neural Networks, and Natural Language Processing. He has published innovative works in top-tier conferences and journals such as IJCAI, EACL, EurJOC and Organic Letters. For more information, please visit https://liuyunqing.github.io/.

\end{IEEEbiography}

\vspace{-24pt}
{\begin{IEEEbiography}[{\includegraphics[width=1in,height=1.25in,clip,keepaspectratio]{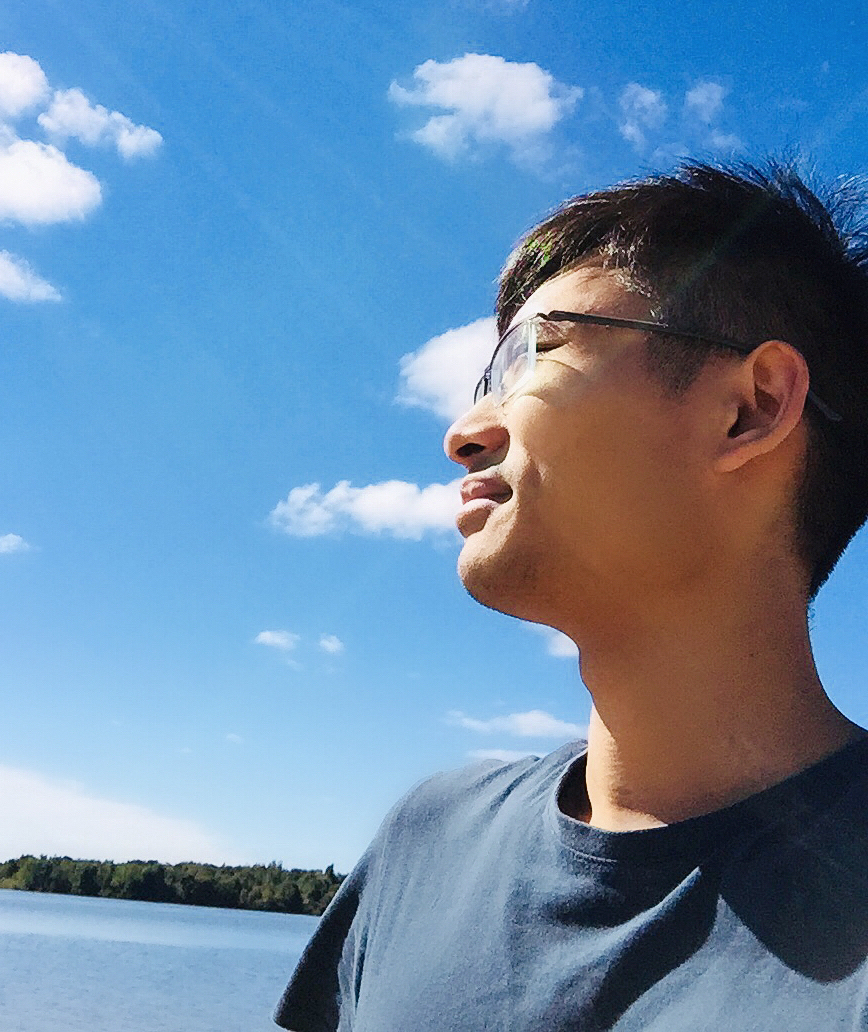}}]{Wenqi Fan} is a research assistant professor of the Department of Computing at The Hong Kong Polytechnic University (PolyU). He received his Ph.D. degree from the City University of Hong Kong (CityU) in 2020.
From 2018 to 2020, he was a visiting research scholar at Michigan State University (MSU). 
His research interests are in the broad areas of machine learning and data mining, with a particular focus on Recommender Systems, Graph Neural Networks, and Trustworthy Recommendations. He has published innovative papers in top-tier journals and conferences such as  TKDE, TIST, KDD, WWW, ICDE, NeurIPS, SIGIR, IJCAI, AAAI, RecSys, WSDM, etc. 
He serves as top-tier conference (senior) program committee members and session chairs (e.g., ICML, ICLR, NeurIPS, KDD, WWW, AAAI, IJCAI, WSDM, etc.), and journal reviewers (e.g., TKDE, TIST, TKDD, TOIS, TAI, etc.). 
More information about him can be found at https://wenqifan03.github.io.

\end{IEEEbiography}

\vspace{-24pt}
{\begin{IEEEbiography}[{\includegraphics[width=1in,height=1.25in,clip,keepaspectratio]{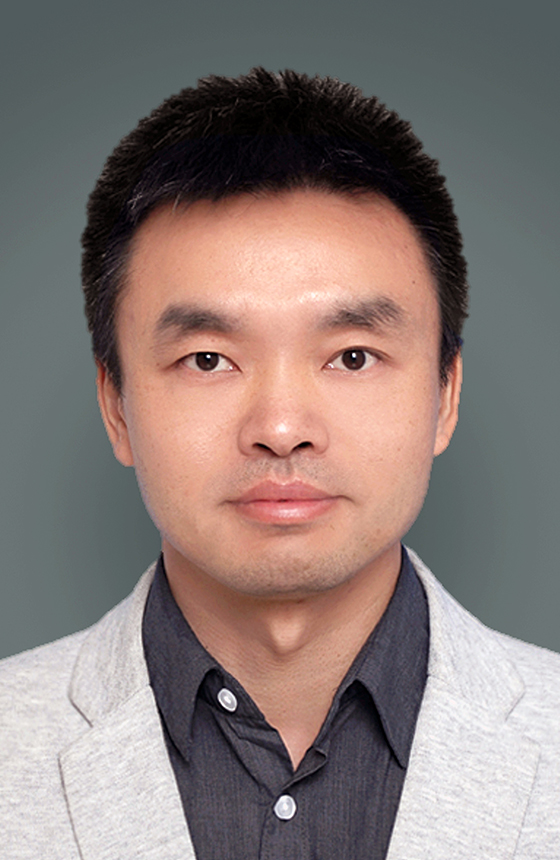}}]{Xiao-yong Wei} is a visiting professor of the Department of Computing, The Hong Kong Polytechnic University. He has been a professor and the head of the Department of Computer Science, Sichuan University of China since 2010. He received his Ph.D. in Computer Science from the City University of Hong Kong and has worked as a postdoctoral fellow in the University of California, Berkeley. His research interests include Multimedia Computing, Health Computing, Machine Learning, and Large-Scale Data Mining. He is a senior member of IEEE, and has served as an associate editor of Interdisciplinary Sciences: Computational Life Sciences since 2020, the program chair of ICMR 2019, ICIMCS 2012, and the technical committee member of over 20 conferences such as ICCV, CVPR, ACM MM, ICME, and ICIP. 

\vspace{-24pt}
\end{IEEEbiography}

{\begin{IEEEbiography}[{\includegraphics[width=1in,height=1.25in,clip,keepaspectratio]{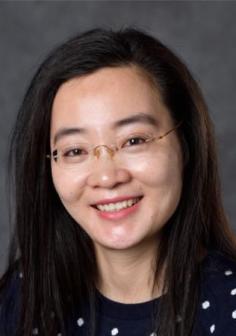}}]{Hui Liu} received her PhD degree in Electrical Engineering from Southern Methodist University in 2015. Her research interests include trustworthy AI, designing data mining algorithms for wireless communication of smart devices, and applying machine learning and data mining in wireless communications. She has more than 7-year research experience in data science on mobile data.  

\end{IEEEbiography}

\vspace{-24pt}
\begin{IEEEbiography}
[{\includegraphics[width=1in,height=1.25in,clip,keepaspectratio]{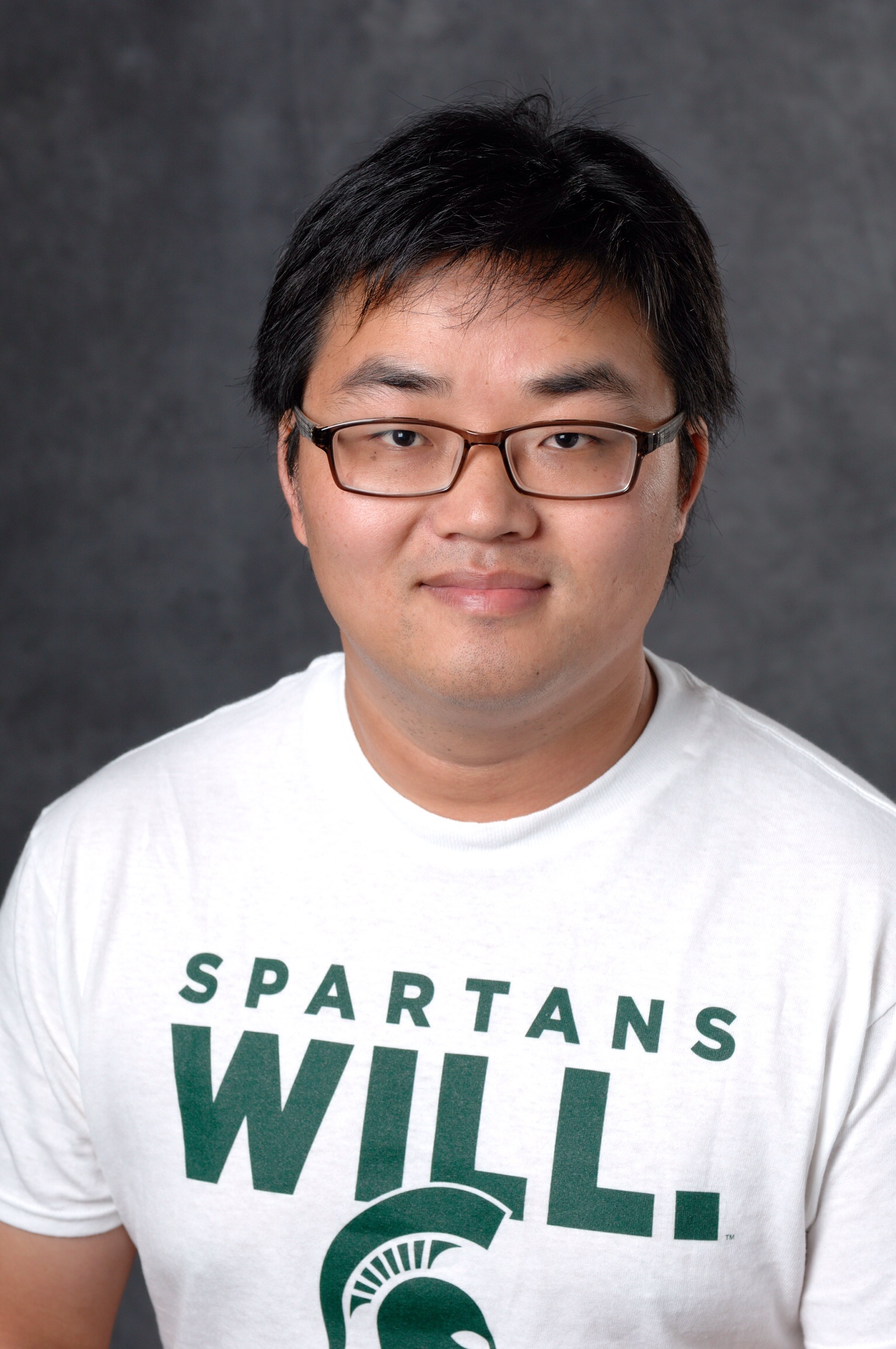}}]{Jiliang Tang} is a University Foundation Professor in the computer science and engineering department at Michigan State University since 2022. He was an associate professor (2021-2022) and an assistant professor (2016-2021) in the same department. Before that, he was a research scientist in Yahoo Research and got his PhD from Arizona State University in 2015 under Dr. Huan Liu. His research interests include graph machine learning, trustworthy AI and their applications in education and biology. He was the recipient of various awards including 2022 AI's 10 to Watch, 2022 IAPR J. K. AGGARWAL Award, 2022 SIAM/IBM Early Career Research Award, 2021 IEEE ICDM Tao Li Award, 2021 IEEE Big Data Security Junior Research Award, 2020 ACM SIGKDD Rising Star Award, 2020 Distinguished Withrow Research Award, 2019 NSF Career Award, and 8 best paper awards (or runner-ups). His dissertation won the 2015 KDD Best Dissertation runner up and Dean's Dissertation Award. He serves as conference organizers (e.g., KDD, SIGIR, WSDM and SDM) and journal editors (e.g., TKDD, TOIS and TKDE). He has published his research in highly ranked journals and top conference proceedings, which have received tens of thousands of citations with h-index 82 (Google Scholar) and extensive media coverage. More details about him can be found at https://www.cse.msu.edu/$\sim$tangjili/.
\end{IEEEbiography}

\vspace{-24pt}
\begin{IEEEbiography}
[{\includegraphics[width=1in,height=1.25in,clip,keepaspectratio]{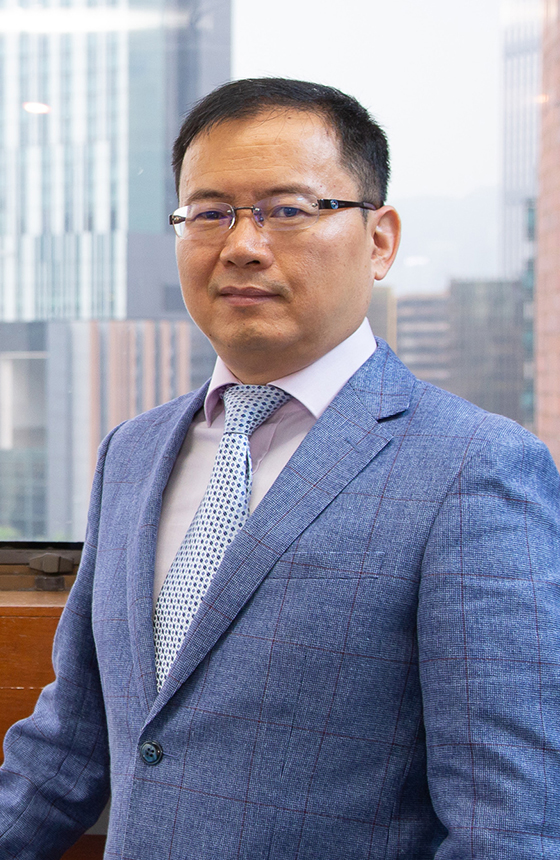}}]{Qing Li}
received the B.Eng. degree from Hunan University, Changsha, China, and the M.Sc. and Ph.D. degrees from the University of Southern California, Los Angeles, all in computer science.
He is currently a Chair Professor (Data Science) and the Head of the Department of Computing, the Hong Kong Polytechnic University. He is a Fellow of IEEE and IET, a member of ACM SIGMOD and IEEE Technical Committee on Data Engineering. 
His research interests include object modeling, multimedia databases, social media, and recommender systems. 
He has been actively involved in the research community by serving as an associate editor and reviewer for technical journals, and as an organizer/co-organizer of numerous international conferences. 
He is the chairperson of the Hong Kong Web Society, and also served/is serving as an executive committee (EXCO) member of IEEE-Hong Kong Computer Chapter and ACM Hong Kong Chapter. In addition, he serves as a councilor of the Database Society of Chinese Computer Federation (CCF), a member of the Big Data Expert Committee of CCF, and is a Steering Committee member of DASFAA, ER, ICWL, UMEDIA, and WISE Society. 
\end{IEEEbiography}
} 


\end{document}